\documentclass[lettersize,journal]{IEEEtran}
\usepackage{amsmath,amsfonts,amssymb}
\usepackage{algorithmic}
\usepackage{algorithm}
\usepackage{array}
\usepackage[caption=false,font=normalsize,labelfont=sf,textfont=sf]{subfig}
\usepackage{textcomp}
\usepackage{stfloats}
\usepackage{url}
\usepackage{verbatim}
\usepackage{graphicx}
\usepackage{cite}
\usepackage{multirow}
\usepackage{xcolor}
\usepackage{booktabs} 
\usepackage{makecell}

\usepackage[colorlinks, citecolor=blue, linkcolor=blue, urlcolor=blue]{hyperref}

\newcommand{\red}[1]{\textcolor{red}{\textbf{#1}}}
\newcommand{\blue}[1]{\textcolor{blue}{\textbf{#1}}}
\newcommand{\yellow}[1]{\textcolor[rgb]{0.96, 0.88, 0.05}{\textbf{#1}}} 

\begin{document}

\title{ASFR-Net: Adversarial Alignment and Spatio-Frequency Refinement Network for Heterogeneous Remote Sensing Image Change Detection}

\author{Xin-Jie Wu,
        Zhi-Hui You,
        Si-Bao Chen,
        Qing-Ling Shu,
        Xiao Wang,
        Jin Tang,
        and Bin Luo
        
\thanks{This work was supported in part by the NSFC Key Project of
Joint Fund for Enterprise Innovation and Development under Grant U24A20342
and in part by the National Natural Science Foundation of China under Grant
62576006 and Grant 61976004. (Xin-Jie Wu and Zhi-Hui You contributed equally
to this work.) (\textit{Corresponding author: Si-Bao Chen.})}%
\thanks{Xin-Jie Wu, Si-Bao Chen, Qing-Ling Shu, Xiao Wang, Jin Tang, and Bin Luo are with the MOE
Key Laboratory of ICSP, IMIS Laboratory of Anhui Province, Anhui Provincial
Key Laboratory of Multimodal Cognitive Computation, Zenmorn-AHU AI Joint
Laboratory, School of Computer Science and Technology, Anhui University,
Hefei 230601, China (e-mail: luoyang\_unique@outlook.com; sbchen@ahu.edu.cn; 2563489133@qq.com; xiaowang@ahu.edu.cn; tangjin@ahu.edu.cn; luobin@ahu.edu.cn).}%

\thanks{Zhi-Hui You is with the School of Public Safety and Emergency Management,
Anhui University of Science and Technology, Hefei 231131, China (e-mail:
youzh@aust.edu.cn).}
}


\maketitle

\begin{abstract}
The core challenge of heterogeneous change detection in remote sensing imagery lies in effectively decoupling genuine land-cover changes from significant modal disparities caused by distinct imaging mechanisms. These intrinsic inconsistencies are prone to introducing pseudo-changes, thereby constraining detection accuracy. To address this, we propose a novel, end-to-end adversarial spatio-frequency refinement network (ASFR-Net). Initially, a modality-invariant representation learner (MIR-Learner) guides the backbone to extract modality-invariant features, effectively bridging the primary domain gap. Subsequently, to address persistent residual modal differences, we design an innovative spatio-frequency synergistic enhancement module (SFEM), which identifies and suppresses sensor-specific noise and artifacts that are difficult to discern in the spatial domain by leveraging frequency-domain processing. Multi-level difference features are then computed from these refined representations and fed into a decoder equipped with cascaded hierarchical guided fusion module (HGFM) blocks to generate precise change maps. To alleviate the data scarcity in heterogeneous tasks, we construct and release a new high-resolution benchmark specifically focused on building changes: the visible-near-infrared heterogeneous change detection (VisNIR-HCD) dataset. It presents unique scientific challenges arising from deceptive visual similarity and non-linear spectral inversions, providing a robust platform for evaluating model generalization. Extensive experiments on VisNIR-HCD and public datasets demonstrate that ASFR-Net achieves state-of-the-art (SOTA) performance, significantly outperforming existing methods. The source code and the VisNIR-HCD dataset are publicly available at \url{https://github.com/LuoYang2024/ASFR-Net}.
\end{abstract}


\section{Introduction}

\IEEEPARstart{C}{hange} {DETECTION} (CD) stands as a fundamental cornerstone in remote sensing image interpretation, dedicated to discerning significant land-cover alterations by analyzing bi-temporal images of the same geographical area acquired at different timestamps \cite{Lv2022_Review}. By delivering timely and precise insights into Earth surface dynamics, CD plays an indispensable role across a spectrum of critical applications, ranging from post-disaster damage assessment \cite{Zheng2021_Disaster} and urban expansion monitoring \cite{Chen2020_STANet} to ecosystem sustainability analysis. In recent years, the explosive proliferation of earth observation programs has led to a surge in high-resolution, multi-source remote sensing data. This data deluge has necessitated a paradigm shift in CD techniques, propelling the field from traditional algebra-based comparisons toward advanced, data-driven deep learning paradigms capable of robust feature representation and high-level semantic abstraction.

Conventionally, most CD methods rely on the assumption of using homogeneous images, where bi-temporal data are acquired by homologous sensors. Under this assumption, unchanged regions exhibit consistent spectral and textural features, allowing algorithms to detect changes based on direct feature distances. However, in practical real-world scenarios, acquiring high-quality homogeneous image pairs is often hindered by adverse weather conditions, cloud cover, or sensor revisit cycles \cite{Wang2025_IA}. To ensure continuous all-weather monitoring, it has become a prevalent trend to utilize multi-source data captured by different sensors, such as combining optical images with synthetic aperture radar (SAR) data, or visible spectrum images with near-infrared (NIR) data \cite{Cheng2024}. This gives rise to the task of heterogeneous CD.

However, compared with homogeneous CD, heterogeneous CD confronts a significantly more intricate and fundamental challenge: modal heterogeneity. Images from different modalities possess distinct imaging mechanisms, which manifest as profound non-linear radiometric differences and disparate feature distributions even for the same ground objects \cite{Sun2022_HCD_Graph}. This implies that the mapping between heterogeneous images is not merely complex but highly non-linear, often involving complex phenomena, such as spectral inversions, that defy any simple linear correlation. Consequently, the direct comparison strategies that are foundational to homogeneous CD are rendered ineffective, as they inherently lack the capacity to distinguish between genuine semantic changes and the spurious differences caused by modal disparity \cite{Jiang2022_HCD}.

Existing heterogeneous CD approaches generally tackle this challenge through two main technical routes: image-to-image translation and feature-level alignment. Translation-based methods \cite{Niu2019,Dong2025_CT2Net} transform heterogeneous images into a pseudo-homogeneous domain. However, the generation process introduces geometric distortions or artifacts, leading to error accumulation in subsequent detection. Feature-level alignment methods, such as AFENet \cite{Pu2024} and HeteCD \cite{Jing2025_XiongAn}, utilize adversarial learning or metric constraints to project features into a shared latent space. Despite the significant progress achieved by these feature-level alignment methods, they typically encounter three intrinsic limitations. First, they face a severe alignment-discriminability trade-off. This aggressive alignment often inadvertently minimizes the distance between changed features in the source domain and background features in the target domain, thereby eroding the semantic discriminability required to accurately detect changes \cite{Jiang2024_CMMAN}. Second, these methods often suffer from the neglect of frequency-domain priors. Most current approaches operate solely in the spatial domain. However, residual modal differences, such as sensor-specific noise patterns or spectral shifts, are often deeply entangled with semantic content in the spatial domain but are readily separable in the frequency domain \cite{Chen2024_FIMP}. Neglecting frequency information restricts the capability of the model to filter out stubbornly persistent modal noise that spatial convolutions struggle to resolve. Third, existing frameworks frequently overlook the semantic-spatial gap and the amplification of cross-modal noise during decoding. While recent homogeneous CD methods employ advanced multi-level aggregation strategies to recover details, applying these directly to heterogeneous tasks often fails. Shallow-level difference features in heterogeneous pairs are highly contaminated by modality-specific noise. Without dedicated selective filtering, existing aggregation modules indiscriminately propagate or even amplify this cross-modal noise into the final prediction. Leveraging aligned deep semantic priors to gate shallow details thus remains a bottleneck for precise boundary delineation. Furthermore, the development of supervised heterogeneous CD methods is constrained by the lack of extensive, high-resolution heterogeneous benchmarks, particularly for visible-NIR scenarios, which are crucial for urban vegetation and building monitoring. This data scarcity fundamentally restricts the training of robust deep models, often resulting in poor generalization when facing complex real-world spectral variations and diverse environmental conditions.

To address these challenges, we propose a novel end-to-end framework named the adversarial spatio-frequency refinement network (ASFR-Net). First, we design a modality-invariant representation learner (MIR-Learner) incorporating a gated adversarial domain unifier (GADU) and a polarity-aware feature regularizer (PAFR). GADU utilizes prediction maps to condition the alignment, ensuring semantic consistency, while PAFR imposes geometric constraints to prevent feature space collapse, preserving discriminability. Second, to address residual modal discrepancies, we introduce a spatio-frequency synergistic enhancement module (SFEM). By transforming features into the Fourier domain, SFEM identifies and suppresses high-frequency modal noise and low-frequency style biases imperceptible in the spatial domain, injecting purified frequency-domain priors back into the network. Finally, to overcome the semantic-spatial gap, a decoder equipped with cascaded hierarchical guided fusion module (HGFM) blocks is employed. HGFM uses a residual dynamic gating mechanism to leverage deep semantic priors to filter cross-modal noise from shallow difference features, preserving critical high-frequency details for precise change maps. Furthermore, the VisNIR-HCD dataset, a high-resolution benchmark for heterogeneous building CD, is constructed and released.

The main contributions of this work are as follows:
\begin{enumerate}

\item We propose a robust heterogeneous CD method named ASFR-Net. By organically integrating global adversarial alignment with fine-grained spatio-frequency refinement, it explicitly decouples modality-induced discrepancies from genuine semantic changes. Comprehensive evaluations across diverse public benchmarks demonstrate that ASFR-Net yields state-of-the-art (SOTA) performance, decisively outperforming current competitive methods.

\item To bridge the primary domain gap, we design MIR-Learner. It reconciles the intrinsic conflict between feature alignment and semantic discriminability via polarity-aware geometric constraints during adversarial training, effectively preventing feature space collapse while ensuring robust cross-modal semantic consistency.

\item To mitigate residual modal discrepancies, we introduce SFEM. This module leverages frequency-domain priors to explicitly decouple and suppress sensor-specific high-frequency noise and low-frequency style biases, while preserving structural integrity and enhancing feature discriminability through dense aggregation.

\item To bridge the semantic-spatial gap during the decoding phase, we propose HGFM. It employs a residual dynamic gating mechanism that utilizes deep semantic priors to actively filter cross-modal noise from shallow features for the precise delineation of change boundaries.

\item We construct and open-source a high-resolution visible-NIR heterogeneous CD dataset named VisNIR-HCD. Focused on building changes, it presents unique scientific challenges arising from deceptive visual similarity and non-linear spectral inversions, thereby providing a rigorous platform for evaluating model generalization.

\end{enumerate}

\section{Related Work}

Existing CD research traditionally explores surface dynamics by assuming data homogeneity between bi-temporal image pairs. Under this premise, homogeneous CD typically dominates the field and operates on the core assumption that bi-temporal data originate from homologous sensors. Unchanged regions therefore exhibit consistent spectral and textural features which allow for direct feature comparison based on distance metrics. Early methods rely heavily on hand-crafted features and algebraic or statistical techniques including change vector analysis \cite{Bovolo2007}, principal component analysis \cite{Deng2008_Urban, wu2021_PCA}, and multivariate alteration detection \cite{Nielsen2007}. However, they are unable to model high-level semantic information and are sensitive to noise and variations in imaging conditions, which fundamentally limits their effectiveness in complex urban scenes and high-resolution environments. Consequently, deep learning serves as the dominant paradigm in contemporary research. Current architectures evolve from simple early-fusion networks \cite{Daudt2018} to weight-sharing Siamese convolutional neural networks \cite{Fang2022_SNUNet} which independently extract deep features to preserve distinct semantic representations. To further expand the receptive field and capture global contexts, sophisticated models increasingly leverage attention mechanisms \cite{Chen2020_STANet} and Transformers \cite{Chen2021_BIT, Bandara2022_ChangeFormer}. Very recently, state space models such as the Mamba architecture \cite{Chen2024_MambaCD} emerge as highly efficient alternatives by offering linear computational complexity while maintaining robust global spatial contextual modeling. Furthermore, to balance computational costs and accuracy, progressive feature aggregation strategies \cite{Li2022_A2Net, You2024_RFANet, Lv2025_HFTN} actively exploit multi-level hierarchical features to recover fine-grained details and suppress pseudo-changes.

Despite these advancements, homogeneous CD methods face strict constraints due to their heavy reliance on spectral and spatial consistency. In practical applications such as disaster assessment \cite{Zheng2021_Disaster}, post-earthquake damage evaluation, and continuous all-weather monitoring \cite{Wang2025_IA}, adverse weather conditions, dense cloud cover, or lengthy sensor revisit cycles frequently hinder the acquisition of high-quality homogeneous image pairs. While visible light spectral images capture rich textural and boundary details, they suffer from severe information degradation under poor visibility. Conversely, SAR provides all-weather capabilities but introduces severe geometric distortions and speckle noise that obscure building structures. Meanwhile, NIR sensors effectively mitigate atmospheric haze and strongly suppress seasonal vegetation interference, but still exhibit massive spectral disparities compared to visible bands. Consequently, using multi-source data to leverage their complementary advantages becomes an inevitable trend to ensure continuous Earth observation \cite{Cheng2024}. However, this transition to heterogeneous CD introduces a profound challenge known as modal heterogeneity. Distinct imaging mechanisms manifest as severe non-linear radiometric differences, structural inconsistencies, and disparate feature distributions. These intrinsic discrepancies render the direct comparison strategies of homogeneous CD ineffective because they lack the capacity to distinguish between genuine semantic changes and spurious differences caused by modal disparity \cite{Sun2022_HCD_Graph, He2023_CMCD}.

To decouple genuine semantic changes from these modality-induced discrepancies, existing heterogeneous CD approaches predominantly explore the paradigms of image-to-image translation and feature-level alignment. Translation-based strategies aim to map the radiometric characteristics of one modality into another and create pseudo-homogeneous pairs via image regression \cite{SUN202216, Sun2022_Regression} or deep generative models such as generative adversarial networks \cite{Niu2019, Zhu2017, Dong2025_CT2Net}. While conceptually intuitive, this indirect approach inherently suffers from the generation process itself. It frequently introduces geometric distortions, semantic shifts, or synthetic artifacts because the generative models struggle to simultaneously preserve modality-specific intrinsic structures and cross-temporal semantic consistency. These generative flaws inevitably propagate to downstream tasks and severely undermine the reliability and precision of the final results \cite{Wang2025_Refined, Jiang2024_CMMAN}.

To circumvent the pitfalls of image translation, current research heavily shifts toward direct feature-level alignment. This paradigm projects high-level representations from disparate domains into a shared and modality-invariant latent space to achieve semantic consistency. Some methods employ metric learning and graph-based models to minimize statistical distances, namely maximum mean discrepancy, between global feature distributions \cite{Sun2022_HCD_Graph, Liu2025_CFRL}. However, relying on fixed distance metrics limits their flexibility when modeling highly complex and non-linear multimodal relationships. Consequently, adversarial learning based domain adaptation approaches gain immense prominence \cite{Pu2024, Jiang2024_CMMAN}. By establishing a min-max game between a feature extractor and a domain discriminator, these methods adaptively compel the network to produce sensor-agnostic representations while retaining essential semantic content.

Despite its powerful alignment capabilities, adversarial learning confronts an intrinsic alignment-discriminability trade-off. Aggressive global distribution alignment often forces changed source regions to align with target background features and erodes the semantic distinctiveness required to accurately identify subtle changes \cite{Zhang2022_Domain}. Furthermore, existing alignment methods predominantly operate solely in the spatial domain and neglect crucial frequency-domain priors. Residual modal discrepancies, including sensor-specific high-frequency noise and low-frequency global style biases, deeply entangle with semantic content spatially but remain readily separable in the frequency domain \cite{Chen2024_FIMP}. Motivated by these critical limitations, the proposed ASFR-Net reconciles this conflict through a progressive refinement strategy. It employs MIR-Learner equipped with geometric constraints to establish primary alignment without feature space collapse. Concurrently, it leverages a spatio-frequency synergistic module to filter out stubborn modal noise. Ultimately, the framework aggregates these purified and multi-scale features to ensure robust change identification and precise boundary delineation.

\begin{table*}[t]
\vspace{-0.3cm}
\centering
\caption{Summary of widely used change detection datasets and the proposed VisNIR-HCD}
\vspace{-0.1cm}
\label{table_1}
\renewcommand{\arraystretch}{1.25} 
\setlength{\tabcolsep}{5pt} 
\begin{tabular}{l c c c c c c c c}
\hline
{Dataset} & 
{{Resolution (m/pixel)}} &
{{Image pairs}} &
{{Image size}} &
{Train/Val/Test} & 
{{$T_1$/$T_2$ type}} &
{{Change Ratio}} &
{{Change Instances}} \\
\hline

\multicolumn{8}{l}{{Homogeneous change detection datasets}} \\
\hline
LEVIR-CD\cite{Chen2020_STANet}  & 0.5 / 0.5   & 637   & $1024 \times 1024$ & 445/64/128 & RGB/RGB  & 4.63\% & 31,333 \\
WHU-CD\cite{WHU_CD2019}       & 0.2 / 0.2   & 1 & $32207 \times 15354$   & 5947/743/744 & RGB/RGB  & 4.33\% & 12,796 \\

\hline
\multicolumn{8}{l}{{Heterogeneous change detection datasets}} \\
\hline
MT-Wuhan\cite{Zhang2022_Domain}   & 10 / 3 & 1   & $11216 \times 13693$ & 552/129/112 & RGB/SAR  & 15.31\% & 4,204 \\
XiongAn\cite{Jing2025_XiongAn}  & 4 / 8 & 2314  & $512 \times 512$   & 1901/413/413\hyperlink{note_xa}{$^*$} & RGB/SAR  & 2.00\% & 16,846 \\

\hline
\multicolumn{8}{l}{{Proposed heterogeneous change detection dataset}} \\
\hline
{\textbf{VisNIR-HCD (Ours)}} & {0.8 / 0.8} & \textbf{8,432} & ${256 \times 256}$ & {5901/839/1692} & \textbf{RGB/NIR}  & \textbf{3.08\%} & \textbf{18,488} \\
\hline
\multicolumn{8}{l}{\scriptsize{\hypertarget{note_xa}{$^*$}Note: Due to discrepancies between the paper description and the released dataset, statistics follow the actual public data where the validation set serves as the test set.}}
\end{tabular}
\end{table*}

\begin{figure*}[!t]
\centering
\includegraphics[width=0.92\textwidth]{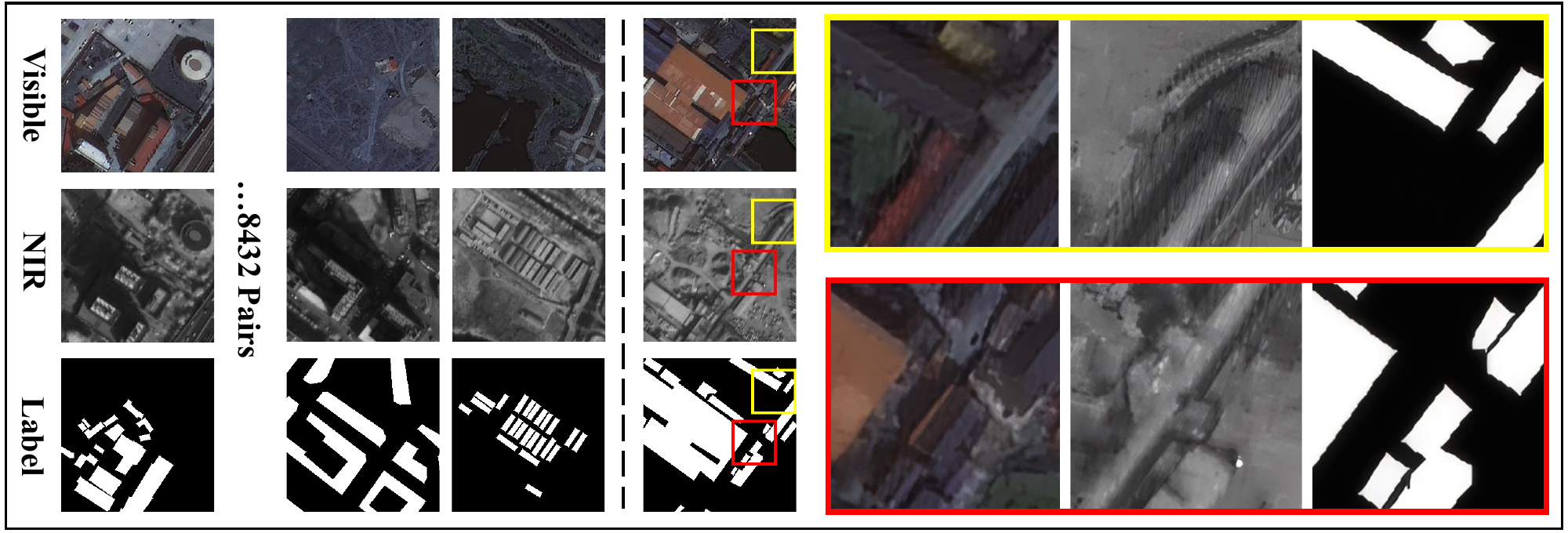}
\vspace{-0.2cm}
\caption{Annotation samples from the proposed VisNIR-HCD dataset.The right side shows close-up of the region with changes in the sample and highlights the challenge of distinguishing buildings due to non-linear spectral differences.}
\label{fig:dataset_vis}
\vspace{-0.2cm}
\end{figure*}

\section{Proposed Benchmark Dataset}
\label{sec:dataset}

In this study, we construct a new benchmark dataset termed VisNIR-HCD to address the lack of high-resolution heterogeneous data. As illustrated in Fig. \ref{fig:dataset_vis}, the dataset encompasses diverse scenes exhibiting significant spectral differences between the visible and NIR modalities. The data collection covers two representative regions in Wuhan, China, spanning $114.04^{\circ}E$--$114.23^{\circ}E$, $30.36^{\circ}N$--$30.61^{\circ}N$ and $114.32^{\circ}E$--$114.53^{\circ}E$, $30.50^{\circ}N$--$30.72^{\circ}N$. Images were acquired by the PMS1 and PMS2 sensors on the Gaofen-2 satellite, featuring diverse landscapes such as urban residential zones, industrial lands, and rural settlements. All images are aligned to the WGS 84 / UTM zone 50N coordinate system, achieving a spatial resolution of 0.8 m/pixel. The dataset uses bi-temporal image pairs that align precisely in space. Each pair consists of a pre-change visible RGB image from 2016 for the pre-change phase $T_1$ and a single-band NIR image from 2023 for the post-change phase $T_2$. To provide a comprehensive statistical profile, the VisNIR-HCD dataset comprises about 552.6 million pixels and a total of 18,488 building change instances. The foreground changed regions account for only 3.08 \% of the total, while the unchanged background dominates at 96.92 \%. This high foreground-to-background ratio reflects the inherent physical sparsity of buildings in geographic space and aligns closely with the distributions observed in other mainstream benchmarks, such as XiongAn \cite{Jing2025_XiongAn} at 2.00\%, WHU-CD \cite{WHU_CD2019} at 4.33\%.

Table \ref{table_1} provides a comprehensive comparison between the proposed VisNIR-HCD and existing mainstream CD datasets. Current benchmarks can be broadly categorized into homogeneous and heterogeneous types. Homogeneous datasets, such as LEVIR-CD and WHU-CD, rely exclusively on RGB imagery. While they offer high spatial resolution, they lack the spectral diversity required to evaluate cross-modal algorithms. Conversely, existing heterogeneous datasets like MT-Wuhan and XiongAn predominantly focus on the optical-SAR modality. While SAR offers critical all-weather observation capabilities, it introduces severe speckle noise, complex multiple scattering effects, and geometric distortions like layover and foreshortening. These factors inherently obscure structural edges and degrade building extraction performance. In contrast, the proposed VisNIR-HCD dataset focuses on the under-explored visible-NIR modality, offering distinct practical benefits. NIR imagery maintains high geometric fidelity and spatial resolution comparable to visible light while providing enhanced atmospheric penetration to mitigate thin clouds or haze. Most importantly, NIR effectively suppresses seasonal vegetation interference, which is a primary source of pseudo-changes in building CD. Because healthy vegetation strongly reflects NIR while man-made structures absorb it, RGB-NIR combinations serve as a highly reliable tool for decoupling seasonal variations from genuine structural changes.

RGB-NIR presents a unique challenge distinct from optical-SAR tasks, namely deceptive similarity. Because RGB and NIR images share high structural consistency and visual clarity, models are prone to erroneously assuming a simple linear mapping between them. However, the true mapping involves complex and category-specific non-linear spectral inversions. Methods tailored for optical-SAR often employ aggressive global distribution alignment to bridge massive modality gaps. When applied directly to RGB-NIR data, such approaches can trigger over-alignment or negative transfer. This inadvertently erases the fine-grained spectral cues essential for distinguishing pseudo-changes from genuine building alterations. 

To ensure both spectral consistency and annotation accuracy, a rigorous multi-stage processing pipeline was implemented. First, the original images underwent radiometric correction and normalization to standardize pixel intensities to the range of $[0, 255]$. Following this pre-processing, a meticulous annotation process was conducted on the large-format and co-registered image pairs. To mitigate visual ambiguities arising from modal heterogeneity, an auxiliary optical reference strategy was employed. By leveraging a corresponding pair of contemporaneous high-resolution optical images as a reference, precise building change labels were generated. These initial labels were then subjected to a dual-person cross-validation protocol, serving as a critical quality control step to filter low-quality samples and ensure high fidelity. Subsequently, these fully annotated and validated large-scale images were partitioned into non-overlapping patches of $256 \times 256$ pixels. The final dataset comprises 8,432 high-quality sample pairs and is randomly partitioned into a training set with 5,901 pairs, a validation set with 839 pairs, and a test set with 1,692 pairs. This strictly follows a 7:1:2 ratio to facilitate a rigorous evaluation of model generalization performance.

Furthermore, to facilitate broader research in homogeneous and heterogeneous CD as well as multi-modal fusion, we simultaneously open-source the complete quad-image sets containing RGB and NIR modalities at both $T_1$ and $T_2$.

\section{The Proposed Approach}
In this section, the overall architecture of the proposed ASFR-Net is first briefly described, and then the intrinsic mechanisms of its constituent modules are elaborated in detail. Finally, the optimization strategy is presented.

\subsection{Approach Overview}
\begin{figure*}[!t]
\vspace{-0.7mm}
\centering
\includegraphics[width=\linewidth]{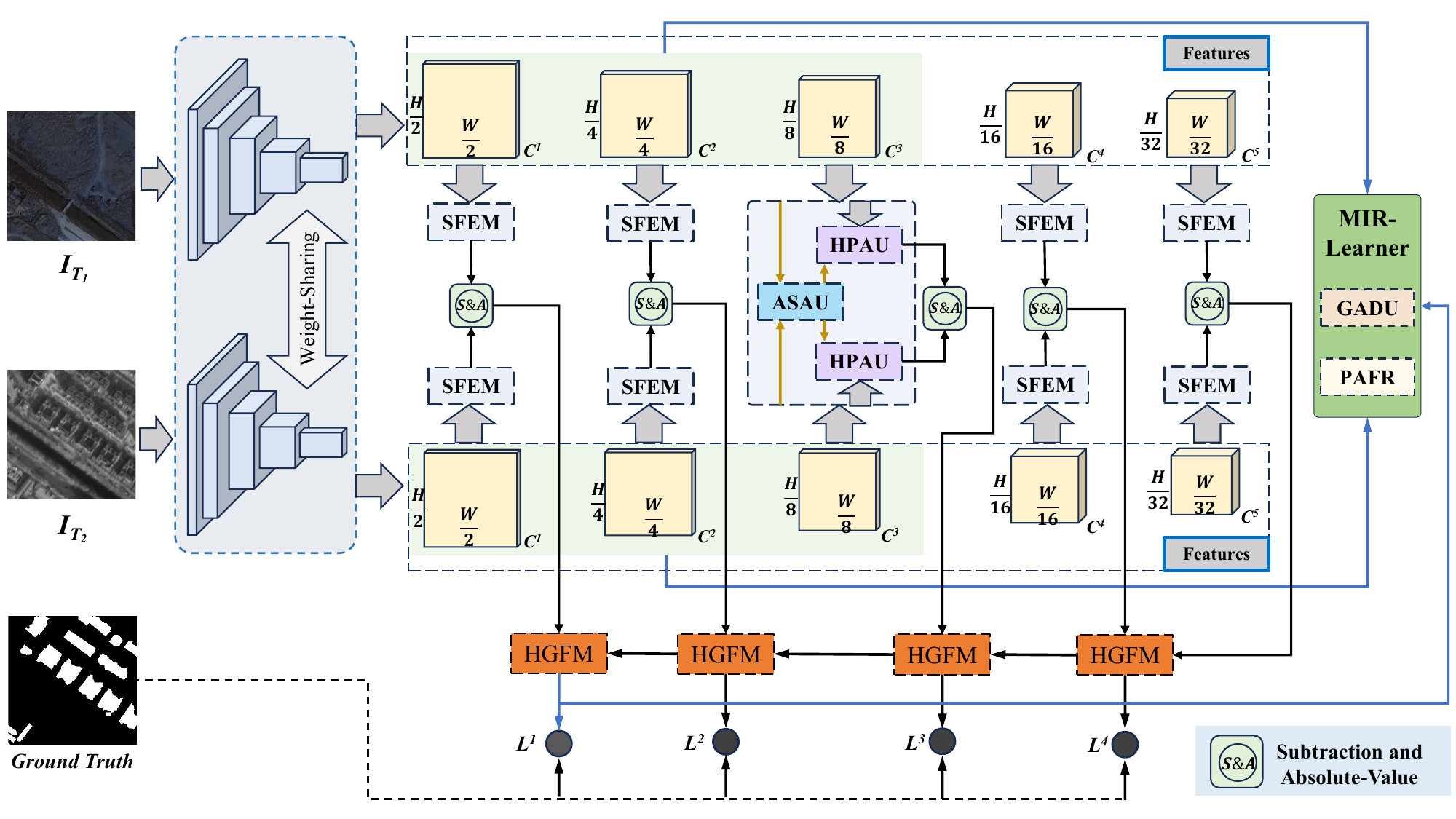}
\caption{Overview of the proposed ASFR-Net. Firstly, a weight-sharing Siamese encoder extracts hierarchical bi-temporal features from the heterogeneous image pair. Then, the MIR-Learner, incorporating the GADU and the PAFR, is utilized to bridge the domain gap and align cross-modal features. Next, the SFEM leverages the ASAU and the HPAU to suppress residual modal noise and refine feature discriminability. Finally, the multi-scale guided fusion decoder employs cascaded HGFM blocks to progressively aggregate multi-level information for precise change map generation.}
\label{fig:architecture}
\vspace{-0.4mm}
\end{figure*}

The schematic architecture of ASFR-Net is illustrated in Fig. \ref{fig:architecture}. Built upon a cascaded feature refinement paradigm, the network synergizes adversarial feature alignment with spatio-frequency enhancement to tackle heterogeneous CD tasks through three key modules: a Siamese encoder guided by a modality-invariant representation learner (MIR-Learner), a spatio-frequency synergistic enhancement module (SFEM), and a decoder equipped with cascaded hierarchical guided fusion module (HGFM) blocks.

\subsubsection{Encoder and Adversarial Alignment}
A weight-sharing Siamese MobileNetV2 \cite{Sandler2018} is employed as the backbone to extract multi-level hierarchical features from a pair of heterogeneous images, $I_{T_1}$ and $I_{T_2}$. We do not assign independent encoders to each modality, as this choice would substantially increase the parameter count of the network, causing the model to overfit sensor-specific noise. In contrast, sharing weights acts as a strict physical regularizer. It compels the network to map disparate inputs into a roughly overlapping latent space from the outset. This shared projection greatly reduces the burden on the subsequent adversarial learning stage; the discriminator only needs to refine a single, unified feature distribution instead of attempting to bridge two isolated spaces. This direct constraint stabilizes gradient updates, enabling the entire framework to converge much faster than a dual-encoder setup. The extracted feature sets are denoted as $\mathcal{F}_{t} = \{F_{t}^1, \dots, F_{t}^5\}$, where $t \in \{T_1, T_2\}$. To bridge the modality gap, the proposed MIR-Learner integrates an adversarial learning mechanism directly into the encoding stage \cite{Ganin2016}. MIR-Learner trains a domain discriminator to identify the modal origin of the features, thereby compelling the encoder to generate modality-invariant representations. While the adversarial pressure from this module aligns global feature distributions, it can inadvertently suppress high-frequency structural details critical for delineating boundaries, necessitating a subsequent feature refinement stage to recover these lost details.

\subsubsection{Spatio-Frequency Synergistic Enhancement}
SFEM is proposed to mitigate the loss of feature discriminability and detail suppression caused by adversarial alignment. The module leverages the insight that structural integrity resides in high-frequency components, which are often attenuated during adversarial training \cite{Jiang2024_CMMAN}. SFEM integrates two key components: an adaptive spectral attention unit (ASAU) that enhances structural features while suppressing modality-specific noise, and a holistic pyramid aggregation unit (HPAU) for spatial aggregation. By injecting the purified frequency priors from ASAU into HPAU, SFEM restores fine-grained structural details. This process yields the refined feature maps $S_t^i$:
\begin{equation}
\label{eq:sfem}
S_{t}^i = \mathrm{}{SFEM}_i(\mathcal{F}_{T_{1}}, \mathcal{F}_{T_{2}}),  t \in \{T_{1}, T_{2}\}, i \in \{1, \dots, 5\},
\end{equation}
where $S_{t}^i$ denotes the structurally enhanced feature at the $i$-th level for the image at time $t$. To generate the change representation, we compute the multi-level difference features:
\begin{equation}
\label{eq:diff}
X^i = \mathrm{Abs}(S_{T_{1}}^i - S_{T_{2}}^i), \quad i \in \{1, \dots, 5\},
\end{equation}
where $\mathrm{}{Abs}(\cdot)$ denotes the element-wise absolute difference operation between the bi-temporal features, and $X^i$ represents the difference feature map at the $i$-th level.

\subsubsection{Decoder}
The decoder comprises cascaded HGFM blocks designed to progressively bridge the gap between global context and local details. To ensure precise object localization and robust suppression of pseudo-changes, each HGFM leverages deep semantic priors to adaptively guide the refinement of shallower, detail-oriented difference features $X^i$:
\begin{equation}
\label{eq:decoder}
P_i,M_i = \mathrm{}{HGFM}(X^i, \mathrm{}{Up}(P_{i+1})), \quad i \in \{4, \dots, 1\},
\end{equation}
where $\mathrm{}{Up}(\cdot)$ denotes the bilinear upsampling operation, and $P_i$ represents the change probability map at the $i$-th level. The final inference result $\textit{M}_{1}$ corresponds to the highest-resolution output $P_1$. A deep supervision strategy is employed during training to impose constraints on predictions across all scales.

\begin{figure}[!t]
\centering
\includegraphics[width=\linewidth]{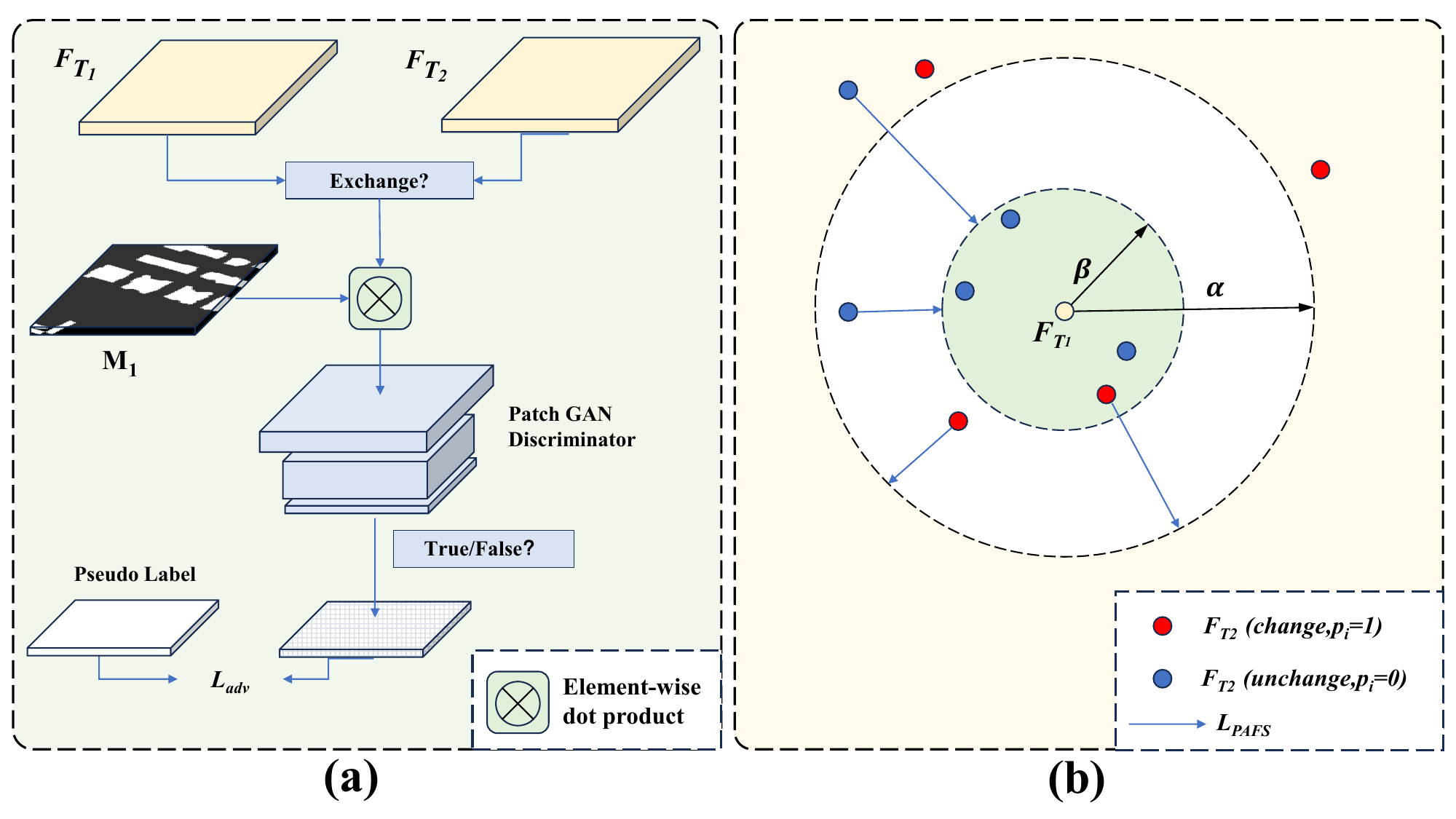}
\vspace{-0.6cm}
\caption{Illustration of the MIR-Learner. (a) GADU: Utilizes ${M}_1$ predicted by the decoder to condition feature alignment via adversarial learning. (b) PAFR: Regularizes feature space geometry by constraining unchanged features within distance $\beta$ and pushing changed features beyond distance $\alpha$.}
\label{fig:mir_learner}
\vspace{-0.2cm}
\end{figure}

\begin{figure}[!t]
\centering
\includegraphics[width=\linewidth]{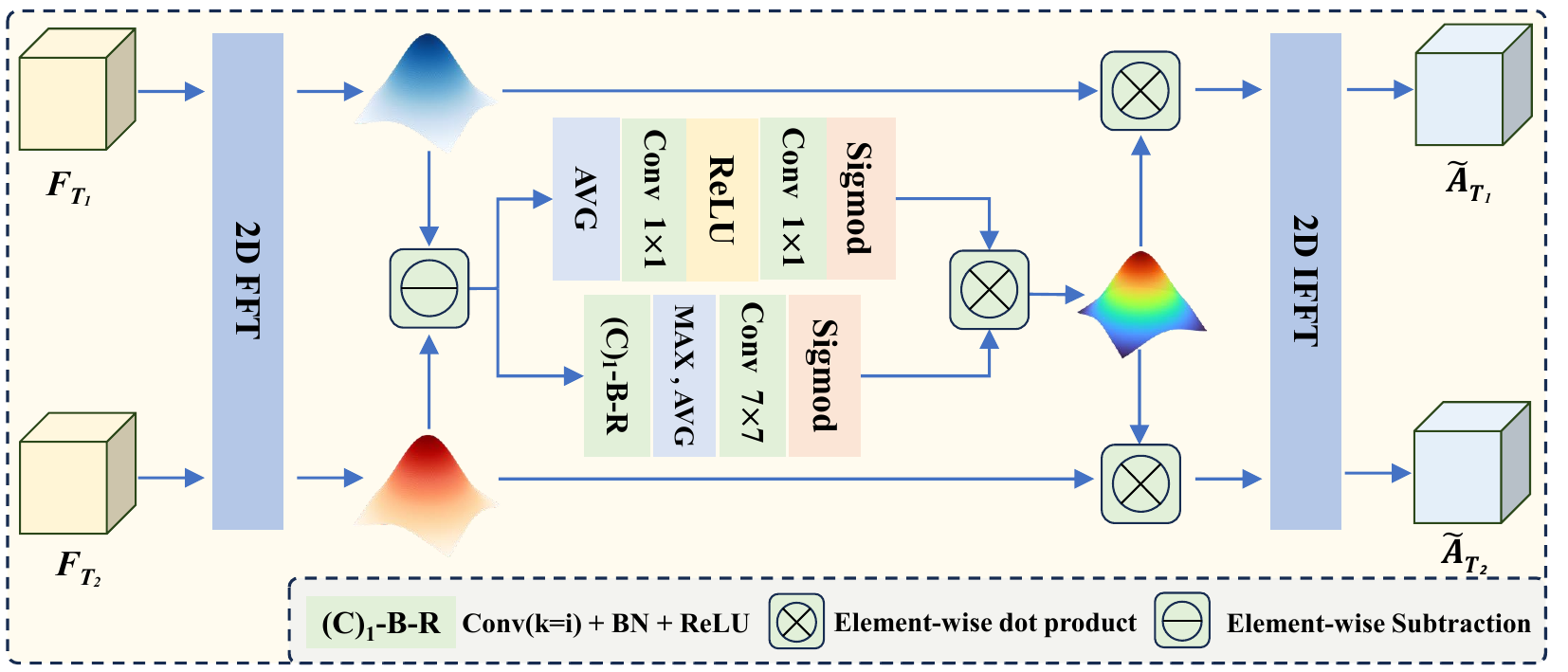}
\caption{Structure of the ASAU. It leverages {FFT} to transform features into the frequency domain, employs a difference-guided attention mask to filter modal noise, and reconstructs the refined features via {IFFT}.}
\label{fig:asau}

\includegraphics[width=\linewidth]{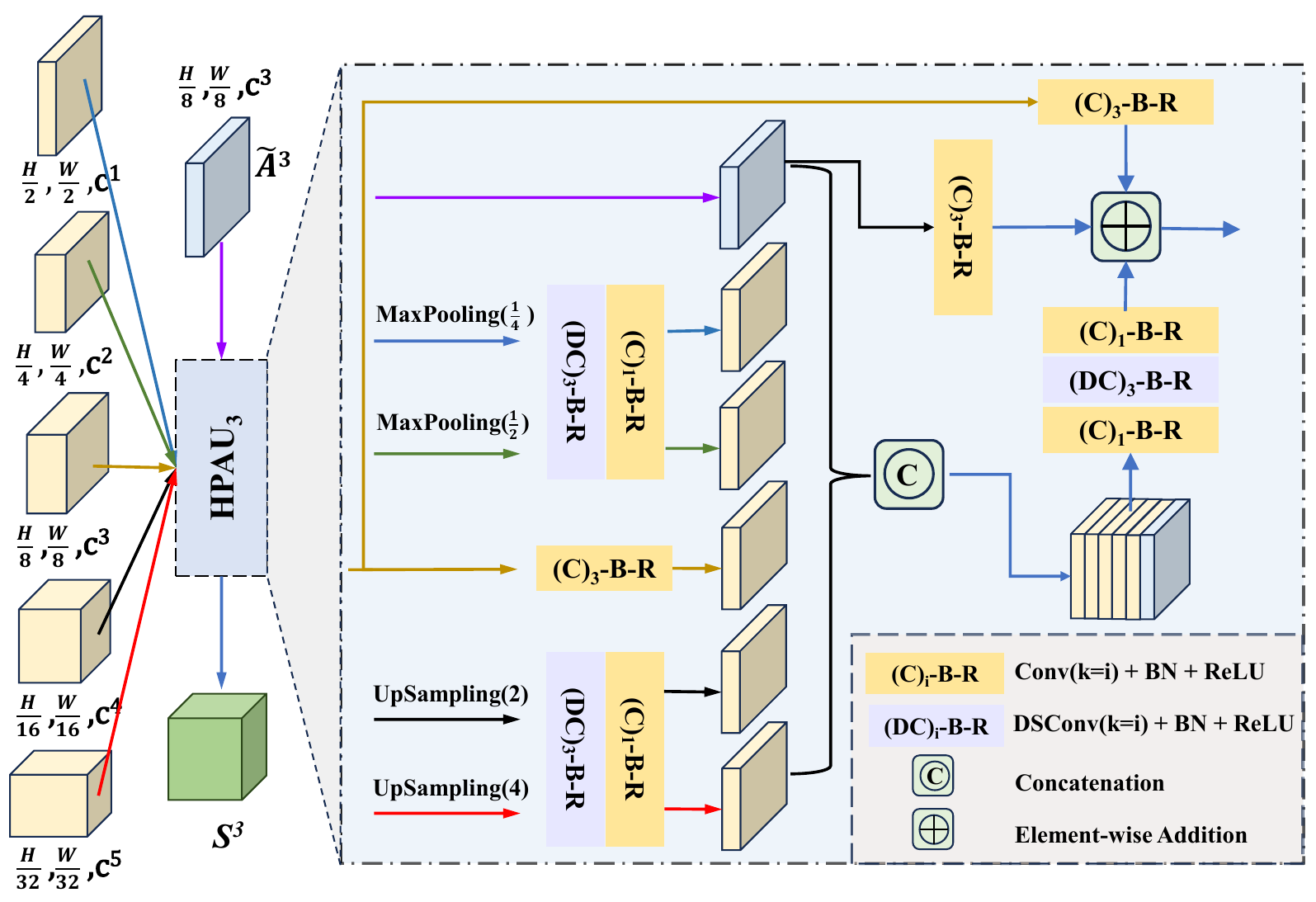}
\vspace{-0.4cm}
\caption{Overview of HPAU, exemplified by the 3rd-level processing pipeline. Leveraging the frequency-refined features from ASAU as guidance, it densely aggregates hierarchical spatial features to reconstruct modality-invariant representations with rich structural details.}
\label{fig:hpau}
\vspace{-0.1cm}
\end{figure}

\subsection{Modality-Invariant Representation Learner (MIR-Learner)}
In heterogeneous CD, a fundamental challenge lies in simultaneously bridging modal disparities and highlighting genuine semantic changes. Since homogeneous CD methods often fail to decouple these signals, MIR-Learner, illustrated in Fig. \ref{fig:mir_learner}, employs an adversarial strategy to disentangle semantic content from modal heterogeneity. This approach ensures that the learned features are invariant to sensor distributions while capturing intrinsic structural changes. It achieves this through two synergistic components: gated adversarial domain unifier (GADU) for distribution alignment and polarity-aware feature regularizer (PAFR) for preserving discriminability.

During the training phase, GADU and PAFR operate strictly in parallel. Multi-scale bi-temporal features extracted by the weight-sharing encoder are simultaneously fed into both modules. They independently evaluate the identical extracted features without modifying them sequentially, thereby forming a complementary push-pull optimization strategy. GADU aggressively aligns global feature distributions to eliminate the macroscopic modality gap, while PAFR operates in parallel to impose explicit geometric boundary constraints, preventing over-alignment and strictly preserving local semantic discriminability. The gradients from their respective objectives are then jointly back-propagated to optimize the shared encoder.

\subsubsection{Gated Adversarial Domain Unifier (GADU)}
The primary objective of GADU is to construct a modality-invariant unified feature space. To address the risk of negative transfer caused by erroneously forcing changed source regions to align with target background regions \cite{Pu2024}, GADU incorporates a prediction-based conditional adversarial mechanism to enforce semantically consistent alignment.

Specifically, we utilize the probability map $P \in \mathbb{R}^{N_c \times H \times W}$ ($N_c=2$) predicted by the decoder to gate the multilinear conditioning on feature map $F \in \mathbb{R}^{C \times H \times W}$. This enables the discriminator to capture the cross-covariance between features and predictions, aligning their joint distributions. The conditional feature $H_c$ is formulated via the outer product:
\begin{equation}
    H_c = F \otimes P \in \mathbb{R}^{(C \times N_c) \times H \times W}.
\end{equation}
This explicitly conditions the multimodal feature distribution on specific semantic categories, ensuring that semantic consistency is preserved throughout the alignment process.

To ensure stability and gradient smoothness, we employ a discriminator $D$ optimized via hinge adversarial loss with an $R_1$ gradient penalty. Unlike standard BCE-based objectives used in AFENet \cite{Pu2024}, this Wasserstein-inspired formulation offers stable gradients and mitigates mode collapse. The specific optimization objective $\mathcal{L}_{D}$ is detailed in Section \ref{sec:optimization}. Conversely, for the generator, the adversarial objective is to minimize the probability of the discriminator correctly identifying target domain features. This is formulated as:
\begin{equation}
    \mathcal{L}_{\text{adv}} = - \mathbb{E}_{H_c \sim \mathcal{T}} [D(H_c)],
\end{equation}
where $\mathcal{T}$ denotes the target domain distribution. By minimizing $\mathcal{L}_{\text{adv}}$, the generator is compelled to extract modality-invariant features that deceive the discriminator, bridging the domain gap between the source ($\mathcal{S}$) and target ($\mathcal{T}$) domains.

\subsubsection{Polarity-Aware Feature Regularizer (PAFR)}

Despite the effectiveness of GADU in domain alignment, it risks over-compressing the feature manifold, potentially compromising semantic discriminability. This trade-off is addressed by PAFR, which enforces geometric regularity and enhances separability in the latent space. Distinct from the divergence-aware contrastive module (DCM) in AFENet \cite{Pu2024}, which primarily targets divergence amplification, PAFR functions as a geometric manifold regularizer designed to explicitly preserve the topological integrity of the feature space.

Formally, PAFR imposes polarity-dependent constraints on the Euclidean distance between $L_2$-normalized bi-temporal features $\mathcal{F}_{T_{1}}$ and $\mathcal{F}_{T_{2}}$. Guided by change labels, we enforce distinct margins: feature pairs in changed regions ($p_i=1$) must separate beyond a lower bound $\alpha$ (repulsion), whereas those in unchanged regions ($p_i=0$) are constrained within an upper bound $\beta$ (attraction). Consequently, the PAFR objective is formulated as a bounded, ReLU-based margin loss:
\begin{equation}
    \mathcal{L}_{\text{PAFR}} = \frac{1}{N} \sum_{i=1}^{N} \left[ p_i \mathrm{ReLU}(\alpha - \delta_i) + (1 - p_i) \mathrm{ReLU}(\delta_i - \beta) \right],
\end{equation}
where $\delta_i = \|F_{T_1}^i - F_{T_2}^i\|_2$ denotes the pixel-wise Euclidean distance. This formulation explicitly penalizes embeddings violating these geometric boundaries, thereby preserving the local manifold structure while enhancing separability. In our implementation, the repulsion margin is set to $\alpha = 1.0$, and the attraction margin is set to $\beta = 0.05$. Since the features are $L_2$-normalized, the maximum possible Euclidean distance between any two vectors is fixed at 2.0. Setting $\alpha = 1.0$ geometrically guarantees an angular separation of at least $60^\circ$ between changed feature pairs, establishing a robust and clear decision boundary in the latent space. Meanwhile, setting $\beta = 0.05$ enforces  the clustering of unchanged features while preventing the network from over-penalizing inherent modality noise. The empirical validation of these boundary settings and their impact on model performance are detailed in the hyperparameter sensitivity analysis in Section \ref{sec:ablation}.

To strike an optimal balance between modality alignment and semantic discriminability, a synergistic optimization strategy is adopted. Instead of treating these modules in isolation, it integrates the multi-scale adversarial loss from GADU and the geometric contrastive loss from PAFR as joint auxiliary regularizers. These terms operate alongside the primary pixel-level CD objective to constitute the total generator loss. The detailed mathematical formulation is provided in Section \ref{sec:optimization}.

\subsection{\hspace{-1.1mm}Spatio-Frequency\hspace{-0.2mm} Synergistic\hspace{-0.15mm} Enhancement\hspace{-0.15mm} Module \hspace{-0.7mm} (SFEM)}
While adversarial alignment mitigates global shifts, residual low-frequency biases and high-frequency artifacts persist. To address this, SFEM employs frequency priors to guide spatial reconstruction, moving beyond spatial-domain-only processing. In our context, modal noise is operationally defined as task-irrelevant, non-semantic spectral components that are strictly induced by disparate sensor mechanisms. This includes high-frequency sensor artifacts, such as SAR speckle or periodic stripes, and low-frequency global biases, such as atmospheric variations or illumination shifts. These components typically lack coherent semantic structure and are precisely what our frequency-domain attention is designed to target. SFEM first filters spectral artifacts from bi-temporal pyramids ($\mathcal{F}_{T_1}, \mathcal{F}_{T_2}$) via ASAU (Fig. \ref{fig:asau}), then performs spectral-guided spatial aggregation with HPAU (Fig. \ref{fig:hpau}).

\subsubsection{Adaptive Spectral Attention Unit (ASAU)}
Pseudo-changes driven by seasonality or illumination exhibit distinct spectral patterns: style biases dominate low frequencies, while noise affects high frequencies. Leveraging this, ASAU adaptively suppresses these task-irrelevant components \cite{Zang2024_FeaSpect}.

The Fourier transform is capable of decoupling style and content, decomposing the previously defined modal noise into specific frequency bands, thus making frequency-domain differencing a reliable tool for heterogeneous CD. To ensure that ASAU primarily addresses these modality-induced discrepancies without removing true change signals, we introduce a dynamic, data-driven difference-guided dual attention mechanism. Unlike traditional frequency ﬁlters that rely on manually ﬁxed cutoff frequencies, our frequency attention gates are adaptively computed based on the speciﬁc content of the spectral difference map between the two modalities. This input-dependent design allows the parallel frequency channel and spatial attention branches to dynamically recalibrate each frequency component. Genuine land-cover changes produce robust, spatially coherent difference signals, which the network learns to adaptively identify and preserve, while simultaneously suppressing modal noise and artifacts. This adaptive decoupling is difficult to achieve in the spatial domain, where style and content are intrinsically entangled, thus validating the reliability and flexibility of our frequency-based approach.

Given spatial features $F_{T_1}^i, F_{T_2}^i \in \mathbb{R}^{C \times H \times W}$, we first perform 2D FFT to compute the spectral difference map $D_{\text{freq}}^i$:
\begin{equation}
    D_{\text{freq}}^i = \mathrm{}{FFT}(F_{T_1}^i) - \mathrm{}{FFT}(F_{T_2}^i) \in \mathbb{R}^{C \times H \times W}.
\end{equation}
By linearity, $D_{\text{freq}}^i$ encapsulates the spectral distribution of differences. To disentangle artifacts, we employ a parallel dual-spectral attention mechanism on the concatenated real and imaginary parts of $D_{\text{freq}}^i$, which we denote as $D_{\text{cat}}^i$.

Specifically, the frequency channel attention (FCA) branch computes the channel gate $G_c^i$ to model inter-channel dependencies. It forms a channel descriptor via global average pooling (GAP) and processes it through a two-layer MLP with sigmoid activation to generate the final gate:
\begin{equation}
    G_c^i = \sigma(\mathrm{}{MLP}(\mathrm{}{GAP}(D_{\text{cat}}^i))),
\end{equation}
where $\sigma(\cdot)$ denotes the sigmoid function. Simultaneously, the frequency spatial attention (FSA) branch localizes noise by applying a large-kernel convolution on channel-pooled features to generate the spatial gate $G_s^i$:
\begin{equation}
    G_s^i = \sigma\left(\mathrm{Conv}_{7 \times 7}\left([\mathrm{}{MaxP}_c(D_{\text{cat}}^i); \mathrm{}{AvgP}_c(D_{\text{cat}}^i)]\right)\right),
\end{equation}
where $\mathrm{}{MaxP}_c$ ($\mathrm{}{AvgP}_c$) denote channel-wise max (average) pooling. The unified gate $G_{\text{total}}^i$ synergistically combines the channel and spatial gates to adaptively recalibrate the features:
\begin{equation}
    G_{\text{total}}^i = G_c^i \odot G_s^i,
\end{equation}
where $\odot$ denotes element-wise multiplication. This unified gate adaptively filters the original spectra to yield purified priors via Inverse FFT:
\begin{equation}
    \tilde{F}_{t}^i = \mathrm{}{IFFT}\left( \mathrm{}{FFT}(F_{t}^i) \odot G_{\text{total}}^i \right), \quad t \in \{T_1, T_2\}.
\end{equation}
This rectifies modal discrepancies in the frequency domain, yielding high-fidelity structural priors for spatial enhancement.

\subsubsection{Holistic Pyramid Aggregation Unit (HPAU)}
Following spectral purification, spatial features often suffer from structural degradation due to aggressive domain alignment. To address this, we propose the HPAU. Unlike standard spatial-only pyramids, HPAU introduces a frequency-guided holistic view that leverages purified priors from ASAU to mitigate modal disparities and restore multi-scale spatial details.

For $j$-th level at time $T_1$, HPAU aggregates the feature hierarchy $\mathcal{F}_{T_{1}}$ guided by the frequency prior $\tilde{F}_{T_1}^j$. First, cross-scale alignment unifies the resolutions of auxiliary features $\{F_{T_1}^k\}_{k \neq j}$ to $j$-th level, yielding aligned contexts $\{F_{T_1}^{j \leftarrow k}\}$. To reconstruct modality-invariant structures, we employ a dual-injection strategy. In the first injection, purified frequency priors are concatenated with spatial contexts to suppress modal noise during dense aggregation:

\begin{equation}
    F_{\text{agg}}^j = \mathrm{}{Conv}_{\text{fuse}}\left( \operatorname{{Concat}}\left[ \{F_{T_1}^{j \leftarrow k}\}_{k \neq j}^5, {F}_{T_1}^j, \tilde{F}_{T_1}^j \right] \right),
\end{equation}
where $\mathrm{}{Conv}_{\text{fuse}}$ denotes the fusion convolution block. Subsequently, to preserve high-frequency details blurred by heterogeneity, the second injection employs a residual mechanism:
\begin{equation}
    S_{T_1}^j =  F_{\text{agg}}^j + \mathrm{}{Conv}_{1 \times 1}(F_{T_1}^j) + \mathrm{}{Conv}_{1 \times 1}(\tilde{F}_{T_1}^j) .
\end{equation}

By synergizing multi-scale semantics with high-fidelity structural priors, HPAU yields a robust, modality-agnostic representation $S_{T_1}^j$ for subsequent processing. Crucially, the residual connection that re-injects the original, unfiltered spatial features serves as a second critical safeguard, preventing the loss of fundamental structures and fine-grained boundary details during frequency filtering.

\begin{figure}[!t]
\centering
\includegraphics[width=\linewidth]{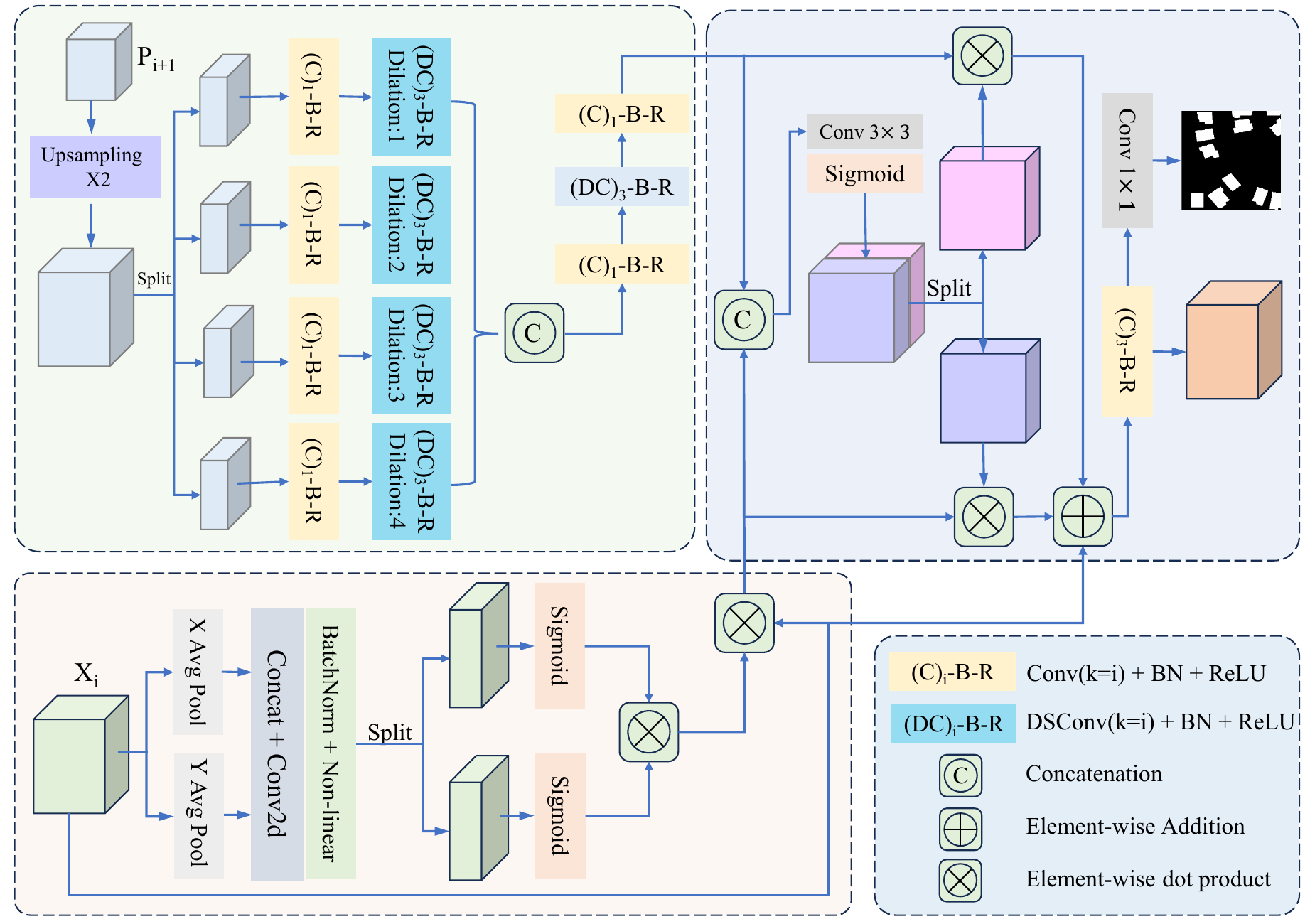}
\caption{Structure of the HGFM. This module uses a residual dynamic gating mechanism to adaptively fuse multi-scale semantic priors from dilated convolutions with spatial details from coordinate attention.}
\label{fig:mgfd}
\vspace{-0.2cm}
\end{figure}

\subsection{Hierarchical Guided Fusion Module (HGFM)}
The decoder reconstructs binary change maps $\{M_i\}_{i=1}^4$ from the SFEM-enhanced difference pyramid $\{X^i\}_{i=1}^5$, where $X^i = |S_{T_1}^i - S_{T_2}^i|$. As highlighted previously, a central bottleneck in heterogeneous CD decoding is bridging the semantic-spatial gap while preventing the propagation of cross-modal noise. Deep-level features possess robust semantics essential for suppressing background interference and pseudo-changes. Conversely, shallow-level difference features contain fine-grained high-frequency details but are highly contaminated by modal disparities. While existing advanced decoders employ various aggregation strategies, they often passively fuse these features and inadvertently propagate cross-modal noise. To reconcile this, we propose the hierarchical guided fusion module (HGFM). Rather than passive aggregation, HGFM is meticulously designed to leverage deep semantic priors to actively filter and selectively guide the refinement of shallow difference features. Initially, the deepest difference feature $X^5$ is processed to generate the coarse-grained semantic prior $P_5$:
\begin{equation}
    P_5 = \mathrm{}{ReLU}(\mathrm{}{BN}(\mathrm{}{Conv}_{3 \times 3}(X^5))),
\end{equation}
where $\mathrm{}{BN(\cdot)}$ denotes batch normalization. Subsequently, an iterative fusion process executes from level $i=4$ down to $1$. At each stage, HGFM refines the difference feature $X^i$ using guidance from the upsampled higher-level feature $\mathrm{}{Up}(P_{i+1})$, yielding the refined output $P_i$ and auxiliary prediction $M_i$. 

HGFM, illustrated in Fig. \ref{fig:mgfd}, serves as the core execution unit designed to seamlessly integrate the shallow detail feature $X^i$ and the deep guidance feature $\mathrm{}{Up}(P_{i+1})$. Unlike simple concatenation, the module operates through a coherent enhance-then-fuse strategy involving two parallel streams.

On the semantic side, to harvest robust multi-scale contextual priors from the guidance feature, we employ a semantic split-aggregation strategy inspired by RFANet \cite{You2024_RFANet}. The input channels are partitioned into multiple groups, each processed by a parallel atrous convolution branch with a distinct dilation rate. This design enables the capture of diverse receptive fields. To efficiently consolidate these features, a depthwise separable fusion block is applied to generate the unified context feature:
\begin{equation}
    F_{\text{ctx}} = \mathrm{}{DSConv}([o_1; \dots; o_G]),
\end{equation}
where $o_g$ denotes the output feature map from the $g$-th parallel atrous convolution branch, and $\mathrm{}{DSConv}(\cdot)$ denotes the depthwise separable convolution for efficient feature fusion.

In parallel, addressing the issue that the shallow feature $X^i$ is detail-rich yet prone to background noise, we utilize coordinate attention \cite{Hou2021}. This mechanism aggregates features along horizontal and vertical directions to generate direction-aware attention maps ($A_h, A_w$), which recalibrate the input to highlight change-relevant spatial structures:
\begin{equation}
    \tilde{X}^{i} = X^i \odot A_h \odot A_w.
\end{equation}

Next, to optimally integrate the semantic context $F_{\text{ctx}}$ and the enhanced spatial details $\tilde{X}^{i}$, we introduce a dynamic gating mechanism. A gate generator $\mathcal{G}$ learns adaptive weight maps $[W_{det}, W_{ctx}]$ based on the concatenated representations. The final refined output $P_i$ is derived through a weighted fusion, reinforced by a critical residual connection to the original signal $X^i$ to preserve high-frequency boundary information:
\begin{equation}
    [W_{det}, W_{ctx}] = \mathrm{}{Softmax}(\mathcal{G}([\tilde{X}^{i}; F_{\text{ctx}}])),
\end{equation}
\begin{equation}
    P_i = \mathrm{}{Conv}_{3 \times 3}((W_{det} \odot \tilde{X}^{i}) + (W_{ctx} \odot F_{\text{ctx}}) + X^i),
\end{equation}
where $\mathrm{}{Conv}_{3 \times 3}$ denotes the final refinement convolution layer. Here, the gate generator $\mathcal{G}$ is implemented as a lightweight $3 \times 3$ convolutional layer that maps the concatenated features to a two-channel logit map, representing the un-normalized importance scores for the detail and context streams.

\subsection{Optimization Objective}
\label{sec:optimization}
The proposed framework is optimized end-to-end via an adversarial  minimax game between the generator $G$ and the domain discriminator $D$. A composite objective is designed to synergistically train the network for  both robust cross-modal feature alignment and high-precision CD.

The generator's primary objective is to produce accurate change maps, guided by a composite loss function $\mathcal{L}_{G}$. The core component is the CD loss $\mathcal{L}_{\text{CD}}$, which employs deep supervision to ensure effective gradient propagation across hierarchical feature levels. Specifically, supervision signals are aggregated from the multi-scale predictions $\{M_i\}_{i=1}^4$. The resulting CD loss $\mathcal{L}_{\text{CD}}$ is formulated as a weighted sum:
\begin{equation}
    \mathcal{L}_{\text{CD}} = \sum_{i=1}^{4} \mathcal{L}_{\text{base}}(M_i, \mathrm{}{Down}_i(Y)),
\end{equation}
where $Y$ is the ground truth, $\mathrm{}{Down}_i(\cdot)$ represents downsampling to the resolution of the $i$-th scale, and the base loss $\mathcal{L}_{\text{base}}$ jointly optimizes binary cross-entropy (BCE) and dice Loss to counteract the inherent class imbalance.

In addition to pixel-level supervision, the generator is constrained by an adversarial loss $\mathcal{L}_{\text{adv}}$ to enforce modality invariance and a geometric contrastive loss $\mathcal{L}_{\text{PAFR}}$ for manifold structure regularization. This synergistic formulation ensures the network learns features that are both aligned across modalities and discriminative for the CD task. Consequently, the total generator objective is formulated as:
\begin{equation}
    \mathcal{L}_{G} = \mathcal{L}_{\text{CD}} + \lambda_{\text{adv}}
    \sum_{k=1}^{K} \mathcal{L}_{\text{adv}}^k + \lambda_{\text{con}}
    \sum_{k=1}^{K} \mathcal{L}_{\text{PAFR}}^k,
\end{equation}
where $\lambda_{\text{adv}} = 0.001$ and $\lambda_{\text{con}} = 0.002$ balance the contributions of the adversarial and geometric regularization losses, respectively. Because both losses are aggregated over multiple dense feature pyramids, their raw gradients are exceptionally large, and their coefficients must be strictly scaled down to prevent these auxiliary gradients from overwhelming the primary pixel-level segmentation objective. Furthermore, a slightly stronger regularization push is required to preserve fine-grained structural boundaries, ensuring they function purely as gentle manifold regularizers, guiding the latent space distribution without disrupting the stable convergence of the primary CD task. A comprehensive ablation study empirically verifying this gradient balancing strategy is provided in Section \ref{sec:ablation}.

Concurrently, the discriminator $D$, comprising multi-scale sub-networks $\{D_k\}_{k=1}^K$, is trained to distinguish between real (source) and fake (target) conditioned feature distributions. To ensure training stability, we employ the Hinge Adversarial Loss augmented with an $R_1$ gradient penalty. The total discriminator objective $\mathcal{L}_{D}$ is aggregated across all scales:
\begin{equation}
\begin{split}
    \mathcal{L}_{D} = \sum_{k=1}^{K} \Big( & \mathbb{E}_{H_{c, \text{real}}^k} [\max(0, 1 - D_k(H_{c, \text{real}}^k))] \\
    + & \mathbb{E}_{H_{c, \text{fake}}^k} [\max(0, 1 + D_k(H_{c, \text{fake}}^k))] \\
    + & \mathbb{E}_{H_{c, \text{real}}^k} [\|\nabla D_k(H_{c, \text{real}}^k)\|_2^2] \Big),
\end{split}
\end{equation}
where $H_{c, \text{real}}^k$ and $H_{c, \text{fake}}^k$ denote the conditioned features from the source and target domains at the $k$-th scale, respectively. The interplay between minimizing $\mathcal{L}_{G}$ and maximizing $\mathcal{L}_{D}$ drives the end-to-end optimization of the network.

\section{Experiments and Analysis}
In this section, comprehensive experiments are conducted to assess the effectiveness and robustness of ASFR-Net. We first detail the experimental setup, including benchmark datasets, implementation details, and evaluation metrics. Subsequently, quantitative and qualitative comparisons are conducted against a suite of representative and advanced methods. Finally, we perform ablation studies to validate the impact of core components, backbone selections, and key hyperparameters.

\subsection{Datasets}
The generalization of our method is evaluated on three representative benchmarks. Spanning visible-NIR and optical-SAR modalities, these datasets possess distinct modal disparities, providing diverse and challenging evaluation scenarios.

\subsubsection{VisNIR-HCD Dataset (Ours)} 
As detailed in Section~\ref{sec:dataset}, this is a high-resolution benchmark comprising 8,432 visible-NIR image pairs focused on building changes. It presents a significant challenge in handling non-linear spectral heterogeneity, particularly the distinct reflectance of vegetation and man-made structures between the RGB and NIR bands. For our experiments, we strictly adhere to the official partition of 5,901 training, 839 validation, and 1,692 test pairs.

\subsubsection{MT-Wuhan Dataset \cite{Zhang2022_Domain}} 
This widely-used optical-SAR benchmark captures land-cover changes resulting from rapid urbanization in Wuhan, China. The dataset is partitioned into 552 training, 129 validation, and 112 test pairs of $256 \times 256$ pixels, where the number of unchanged pixels is approximately five times that of changed pixels. Despite the limited sample size, its complex urban textures present a rigorous test for model generalization under data scarcity conditions.

\subsubsection{XiongAn Dataset \cite{Jing2025_XiongAn}} 
This dataset focuses on building CD during large-scale urban redevelopment in Xiong’an New Area, China, comprising GaoFen-2 multispectral and GaoFen-3 SAR images. The inherent speckle noise and geometric distortions of SAR imagery pose severe challenges to detection accuracy. As the official test set is not publicly available, we follow common practice and use the validation set for evaluation. The original images have a resolution of $512 \times 512$ pixels. In our experiments, we crop them into non-overlapping $256 \times 256$ patches, ultimately resulting in 7,604 pairs for training and 1,652 pairs for validation and testing, respectively.

\begin{table*}[!t]
\vspace{-0.2cm}
\caption{Quantitative results of different comparison methods on the VisNIR-HCD, MT-Wuhan, and XiongAn datasets. The evaluation metrics include model complexity [Params (M) and FLOPs (G)] and detection accuracy [Precision (Pre), Recall (Rec), F1-score (F1), Intersection over Union (IoU), and Overall Accuracy (OA)]. Color convention: \red{best}, \blue{2nd-best}, and \yellow{3rd-best}.}
\label{tab:sota_comparison}
\centering
\renewcommand{\arraystretch}{1.2}
\setlength{\tabcolsep}{3pt} 

\begin{tabular}{l|@{\hspace{1.5pt}}c@{\hspace{2.5pt}}c@{\hspace{1.5pt}}|ccccc|ccccc|ccccc}
\hline
\multirow{2}{*}{{Methods}} & \multirow{2}{*}{{\makecell{Params\\(M)$\downarrow$}}} & \multirow{2}{*}{{\makecell{FLOPs\\(G)$\downarrow$}}} & \multicolumn{5}{c|}{{VisNIR-HCD (\%)}} & \multicolumn{5}{c|}{{MT-Wuhan (\%)}} & \multicolumn{5}{c}{{XiongAn (\%)}} \\
\cline{4-18} 
 & & & \makebox[2.4em]{Pre $\uparrow$} & \makebox[2.4em]{Rec $\uparrow$} & \makebox[2.4em]{F1 $\uparrow$} & \makebox[2.4em]{IoU $\uparrow$} & \makebox[2.4em]{OA $\uparrow$} & \makebox[2.4em]{Pre $\uparrow$} & \makebox[2.4em]{Rec $\uparrow$} & \makebox[2.4em]{F1 $\uparrow$} & \makebox[2.4em]{IoU $\uparrow$} & \makebox[2.4em]{OA $\uparrow$} & \makebox[2.4em]{Pre $\uparrow$} & \makebox[2.4em]{Rec $\uparrow$} & \makebox[2.4em]{F1 $\uparrow$} & \makebox[2.4em]{IoU $\uparrow$} & \makebox[2.4em]{OA $\uparrow$} \\
\hline
A2Net \cite{Li2022_A2Net} & 3.78 & 3.05 & \red{82.54} & 75.22 & \yellow{78.71} & \yellow{64.89} & \blue{98.77} & 56.89 & 52.81 & 54.78 & 37.72 & 87.80 & 84.82 & 82.19 & \yellow{83.49} & \yellow{71.65} & 98.99 \\
RFANet \cite{You2024_RFANet} & 2.86 & 3.16 & 82.17 & \yellow{75.30} & {78.58} & {64.72} & \yellow{98.76} & 56.74 & 58.82 & 57.76 & 40.61 & 87.97 & 82.06 & \blue{83.83} & 82.94 & 70.85 & 98.93 \\
STADE-CDNet \cite{Li2024_STADE} & 3.50 & 11.99 & 74.84 & 51.49 & 61.01 & 43.90 & 98.02 & 32.62 & 53.80 & 40.62 & 25.49 & 78.00 & 56.57 & 53.80 & 55.15 & 38.07 & 97.28 \\
MambaCD \cite{Chen2024_MambaCD} & 85.53 & 44.83 & 76.43 & 74.24 & 75.32 & 60.41 & 98.53 & 58.50 & 54.63 & 56.50 & 39.37 & 88.24 & 79.20 & \red{86.20} & 82.55 & 70.29 & 98.87 \\
STENet \cite{Pan2024_STENet} & 10.95 & 14.73 & 80.53 & 68.67 & 74.13 & 58.89 & 98.56 & 42.93 & 51.66 & 46.89 & 30.63 & 83.64 & 81.66 & 80.01 & 80.83 & 67.83 & 98.82 \\
DGMA2-Net \cite{Ying2024_DGMA2} & 37.10 & 18.10 & 80.81 & 73.39 & 76.92 & 62.50 & 98.67 & 52.27 & \blue{66.97} & \yellow{58.72} & \yellow{41.56} & 86.83 & 83.05 & 82.30 & 82.67 & 70.46 & 98.93 \\
CSI-Net \cite{Liu2024_CSI} & 62.18 & 367.25 & 75.63 & 59.18 & 66.40 & 49.70 & 98.20 & 43.83 & 49.95 & 46.69 & 30.46 & 84.04 & 79.53 & 69.12 & 73.96 & 58.68 & 98.57 \\
AFENet \cite{Pu2024} & 39.7 & 398.58 & 68.23 & 68.79 & 68.51 & 52.11 & 98.09 & 41.79 & \red{69.26} & 52.13 & 35.25 & 82.21 & 81.18 & 80.92 & 81.05 & 68.14 & 98.82 \\
CASP \cite{Wang2025_CASP} & 1.74 & 2.63 & 81.28 & 72.16 & 76.43 & 61.86 & 98.66 & \red{63.48} & 47.90 & 54.60 & 37.55 & \blue{88.86} & \blue{86.78} & 80.30 & 83.41 & 71.55 & \blue{99.01} \\
ConvFormer-CD \cite{Yang2025_ConvFormer} & 37.72 & 5.14 & 69.44 & 67.38 & 68.40 & 51.97 & 98.10 & \blue{62.43} & 45.77 & 52.81 & 35.88 & 88.52 & 84.03 & 79.46 & 81.68 & 69.03 & 98.89 \\
SFEARNet \cite{Li2025_SFEARNet} & 5.56 & 4.65 & 78.89 & 75.09 & 76.94 & 62.53 & 98.64 & 54.61 & 61.22 & 57.73 & 40.57 & 87.46 & \yellow{85.76} & 80.39 & 82.99 & 70.93 & 98.97 \\
EFICNN \cite{Liu2025_EFICNN} & 21.32 & 91.34 & 81.93 & \blue{75.87} & \blue{78.78} & \blue{64.99} & \blue{98.77} & 60.16 & 57.74 & \blue{58.92} & \blue{41.77} & \yellow{88.74} & 85.43 & 81.14 & 83.23 & 71.28 & 98.98 \\
HeteCD \cite{Jing2025_XiongAn} & 63.85 & 81.28 & 75.61 & 74.05 & 74.82 & 59.77 & 98.50 & \yellow{61.58} & 55.45 & 58.35 & 41.19 & \red{88.93} & 85.05 & 80.32 & 82.62 & 70.38 & 98.95 \\
HRMNet \cite{Li2025_HRMNet} & 13.46 & 12.17 & \yellow{82.41} & 74.08 & 78.02 & 63.96 & 98.74 & 52.65 & 55.07 & 53.83 & 36.83 & 86.79 & 84.74 & \yellow{82.80} & \blue{83.76} & \blue{72.06} & \yellow{99.00} \\
\hline
{Ours} & 6.35 & 15.13\hyperlink{note_flops}{$\dagger$} & \blue{82.45} & \red{78.53} & \red{80.44} & \red{67.28} & \red{98.86} & 58.68 & \yellow{62.99} & \red{60.76} & \red{43.64} & 88.62 & \red{87.17} & 82.66 & \red{84.86} & \red{73.71} & \red{99.08}\\
\hline
\end{tabular}

\vspace{4pt}
\raggedright
\scriptsize{\hypertarget{note_flops}{$\dagger$}Note: Since standard profiling tools often fail to track hardware-level FLOPs for frequency-domain operations, for methods incorporating ASAU, we manually calculated this overhead based on the Cooley-Tukey algorithm: $\mathcal{O}(C \cdot HW \log_2(HW))$. By meticulously accumulating this across all 5 pyramid levels in our Siamese architecture, the FFT/iFFT operations contribute approximately 0.39 GFLOPs, which has been added to the FLOPs of ASFR-Net to ensure transparency of complexity reporting.}

\vspace{-0.2cm}
\end{table*}

\subsection{Implementation Details and Evaluation Metrics}

Our approach is implemented in PyTorch, with all experiments conducted on a workstation equipped with two NVIDIA GeForce RTX 3090 GPUs (24 GB). We adopt a lightweight MobileNetV2 \cite{Sandler2018} backbone pre-trained on ImageNet as the weight-sharing feature extractor, while other components are randomly initialized. The network is optimized using the AdamW optimizer with $\beta_1=0.9$, $\beta_2=0.99$, and a weight decay of $1e-2$. The initial learning rate is set to 0.0005 with a batch size of 16. A Poly learning rate schedule with a power of 0.9 is employed throughout training. The model is trained for 40k, 2k, and 20k iterations on the VisNIR-HCD, MT-Wuhan, and XiongAn datasets, respectively. A variety of online data augmentation techniques, including random re-scaling, cropping, flipping, temporal exchange, and cutmix, are incorporated to enhance generalization. The model is evaluated on the validation set after each epoch, saving the checkpoint with the highest F1 for final testing.

In line with standard practices, five evaluation metrics are employed to quantify performance: precision (Pre), recall (Rec), f1-score (F1), intersection over union (IoU), and overall accuracy (OA). Among these, F1 and IoU are considered the primary indicators for comprehensive evaluation.

\subsection{Quantitative Analysis}

\begin{figure*}[!t]
\vspace{-0.2cm}
\centering
\includegraphics[width=\textwidth]{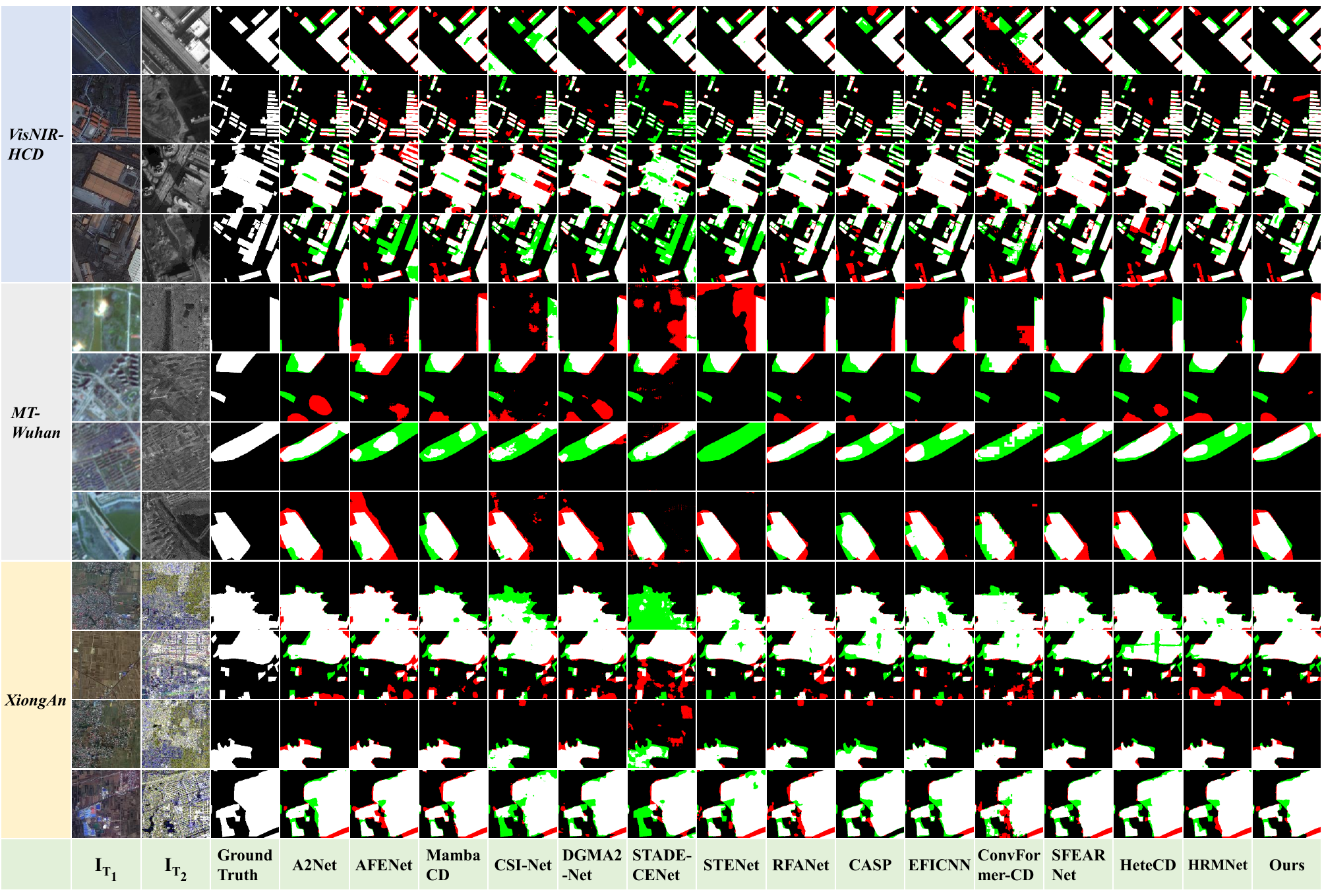}
\vspace{-0.6cm}
\caption{Qualitative visualizations of different methods tested on VisNIR-HCD, MT-Wuhan and XiongAn datasets. The visualizations employ a standard color scheme for error analysis: true positives (TP) in white, true negatives (TN) in black, false positives (FP) in red, and false negatives (FN) in green. }
\label{fig:visual_results}
\vspace{-0.2cm}
\end{figure*}

Table \ref{tab:sota_comparison} details the quantitative comparison results of ASFR-Net against related  methods, including A2Net \cite{Li2022_A2Net}, RFANet \cite{You2024_RFANet}, STADE-CDNet \cite{Li2024_STADE}, MambaCD \cite{Chen2024_MambaCD}, STENet \cite{Pan2024_STENet}, DGMA2-Net \cite{Ying2024_DGMA2}, CSI-Net \cite{Liu2024_CSI}, AFENet \cite{Pu2024}, CASP \cite{Wang2025_CASP}, ConvFormer-CD \cite{Yang2025_ConvFormer}, SFEARNet \cite{Li2025_SFEARNet}, EFICNN \cite{Liu2025_EFICNN}, HeteCD \cite{Jing2025_XiongAn}, and HRMNet \cite{Li2025_HRMNet}, on the three datasets. As evidently observed, ASFR-Net consistently achieves superior performance across key metrics, demonstrating its exceptional robustness in handling diverse heterogeneous scenarios.

In the challenging VisNIR-HCD dataset where non-linear spectral discrepancies between visible and NIR bands often induce significant pseudo-changes, ASFR-Net secures a leading F1 of 80.44\% and an IoU of 67.28\%. This performance substantially surpasses traditional CNN-based Siamese networks such as RFANet \cite{You2024_RFANet} and A2Net \cite{Li2022_A2Net}. It is worth noting that while the registration-aware method CASP \cite{Wang2025_CASP} performs competitively on other datasets, its performance drops on VisNIR-HCD with an F1 of 76.43\%. This indicates that relying solely on spatial alignment is insufficient to bridge the profound visible-NIR spectral gap. Furthermore, compared to the heterogeneous-specific method AFENet \cite{Pu2024} which yields an F1 of 68.51\%, our approach achieves a remarkable improvement of nearly 12\%. This significant margin suggests that the global adversarial strategy employed by AFENet tends to over-smooth high-frequency details, whereas our coarse-to-fine strategy effectively preserves critical boundary information while mitigating spectral inconsistencies through the frequency-domain refinement module.

Extending the evaluation to optical-SAR scenarios, ASFR-Net exhibits exceptional resistance to inherent speckle noise and geometric distortions found in the MT-Wuhan and XiongAn datasets. On the noise-heavy MT-Wuhan benchmark, our method outperforms top competitors like EFICNN \cite{Liu2025_EFICNN} and DGMA2-Net \cite{Ying2024_DGMA2} by approximately 2\% in terms of F1. Furthermore, it is worth noting that although lightweight models such as A2Net \cite{Li2022_A2Net} establish a competitive baseline on VisNIR-HCD by exploiting spatial structural similarity, relying solely on spatial alignment is a fragile strategy for heterogeneous tasks. This limitation becomes extremely evident when evaluated on the MT-Wuhan dataset; due to extreme speckle noise and geometric distortions, the performance of A2Net drops sharply, yielding an F1 score of only 54.78\%. In stark contrast, ASFR-Net exhibits outstanding robustness, outperforming A2Net by roughly 6\% in both F1 and IoU. ASFR-Net strikes a better balance by leveraging SFEM to isolate and filter out sensor-specific noise components in the frequency domain, thereby effectively suppressing background false alarms that spatial-only methods often struggle to decouple.

Beyond raw detection accuracy, computational efficiency is a critical factor for practical deployment in real-world remote sensing applications. As detailed in Table \ref{tab:sota_comparison}, many existing high-performance models come with prohibitive computational costs. For instance, MambaCD \cite{Chen2024_MambaCD} and AFENet \cite{Pu2024} require massive parameters exceeding 85 M and 39 M respectively, along with high FLOPs, which severely hinders their deployability on edge devices. Conversely, ASFR-Net adopts a relatively efficient design with only 6.35 M parameters and 15.13 G FLOPs. Although our model incurs a marginal increase in computational overhead compared to existing ultra-lightweight networks like RFANet \cite{You2024_RFANet}, this trade-off is highly justified by the significant performance gains. Specifically, ASFR-Net outperforms RFANet by a solid margin of 2.56\% in terms of IoU on the VisNIR-HCD dataset. Consequently, ASFR-Net establishes an optimal equilibrium between model complexity and detection accuracy, ensuring high-precision inference while maintaining a manageable memory footprint suitable for large-scale data processing.

\begin{table*}[!t]
\vspace{-0.3cm}
\caption{Ablation study for different components on the three datasets. Color convention: \red{best}, \blue{2nd-best}, and \yellow{3rd-best}.}
\label{tab:ablation_study_1}
\centering
\renewcommand{\arraystretch}{1.2} 
\setlength{\tabcolsep}{1.95pt} 
\begin{tabular}{l|@{\hspace{1.5pt}}c@{\hspace{2.5pt}}c@{\hspace{1.5pt}}|ccccc|ccccc|ccccc}
\hline
\multirow{2}{*}{{Methods}} & \multirow{2}{*}{{\makecell{Params\\(M)$\downarrow$}}} & \multirow{2}{*}{{\makecell{FLOPs\\(G)$\downarrow$}}} & \multicolumn{5}{c|}{{VisNIR-HCD (\%)}} & \multicolumn{5}{c|}{{MT-Wuhan (\%)}} & \multicolumn{5}{c}{{XiongAn (\%)}} \\
\cline{4-18} 
 & & & \makebox[2.4em]{Pre $\uparrow$} & \makebox[2.4em]{Rec $\uparrow$} & \makebox[2.4em]{F1 $\uparrow$} & \makebox[2.4em]{IoU $\uparrow$} & \makebox[2.4em]{OA $\uparrow$} & \makebox[2.4em]{Pre $\uparrow$} & \makebox[2.4em]{Rec $\uparrow$} & \makebox[2.4em]{F1 $\uparrow$} & \makebox[2.4em]{IoU $\uparrow$} & \makebox[2.4em]{OA $\uparrow$} & \makebox[2.4em]{Pre $\uparrow$} & \makebox[2.4em]{Rec $\uparrow$} & \makebox[2.4em]{F1 $\uparrow$} & \makebox[2.4em]{IoU $\uparrow$} & \makebox[2.4em]{OA $\uparrow$} \\
\hline
(a) Baseline & 2.63 & 2.54 & 81.75 & 73.84 & 77.59 & 63.39 & 98.72 & 53.01 & 58.23 & 55.50 & 38.41 & 86.94 & 82.92 & \yellow{84.35} & 83.63 & 71.87 & 98.97 \\
(b) + MIR-Learner & 4.50 & 3.65 & \blue{85.16} & 74.35 & 79.38 & 65.82 & \blue{98.84} & \red{62.29} & 52.94 & 57.23 & 40.09 & \blue{88.94} & 85.13 & 82.69 & 83.89 & 72.25 & 99.01 \\
(c) + MIR-Learner (w/o PAFR) & 4.50 & 3.65 & 81.42 & 76.59 & 78.93 & 65.20 & 98.77 & 57.73 & 55.91 & 56.80 & 39.67 & 88.11 & \blue{87.12} & 80.58 & 83.72 & 72.00 & 99.03 \\
(d) + SFEM & 3.77 & 10.92 & 82.47 & 76.46 & 79.35 & 65.77 & \yellow{98.80} & 52.67 & \red{63.26} & 57.48 & 40.33 & 86.91 & 83.91 & \blue{84.75} & 84.33 & 72.91 & 99.02 \\
(e) + SFEM (w/o HPAU) & 2.68 & 3.11 & 80.00 & 76.41 & 78.16 & 64.16 & 98.71 & 55.70 & 58.86 & 57.23 & 40.09 & 87.70 & 82.81 & \red{85.67} & 84.21 & 72.73 & 99.00 \\
(f) + SFEM (w/o ASAU) & 3.49 & 8.49 & 81.86 & 76.62 & 79.15 & 65.50 & 98.78 & 57.42 & 57.07 & 57.24 & 40.10 & 88.08 & \yellow{85.97} & 82.32 & 84.11 & 72.57 & 99.03 \\
(g) + HGFM & 2.69 & 2.85 & 81.17 & \yellow{77.91} & 79.50 & 65.98 & \blue{98.84} & 53.97 & \yellow{59.49} & 56.60 & 39.47 & 87.24 & \blue{87.12} & 81.26 & 84.09 & 72.54 & \yellow{99.04} \\
(h) + MIR-Learner+SFEM & 5.65 & 12.03 & \red{85.91} & 74.37 & 79.72 & 66.28 & \red{98.86} & 57.11 & 58.04 & 58.08 & 40.92 & 88.28 & 85.22 & 83.94 & \blue{84.57} & \blue{73.27} & \blue{99.05} \\
(i) + MIR-Learner+HGFM & 4.56 & 3.96 & \yellow{84.60} & 75.88 & \yellow{80.00} & \yellow{66.67} & \red{98.86} & \yellow{60.92} & 57.10 & \yellow{58.95} & \yellow{41.80} & \yellow{88.88} & 85.55 & 83.37 & 84.45 & 73.08 & \yellow{99.04} \\
(j) + SFEM +HGFM & 3.83 & 11.23 & 82.32 & \blue{78.27} & \blue{80.24} & \blue{67.01} & \blue{98.84} & \blue{62.21} & 56.77 & \blue{59.37} & \blue{42.21} & \red{89.13} & 85.56 & 83.55 & \yellow{84.55} & \yellow{73.23} & \blue{99.05} \\
(k) ASFR-Net (Ours) & 6.35 & 15.13 & 82.45 & \red{78.53} & \red{80.44} & \red{67.28} & \red{98.86} & 58.68 & \blue{62.99} & \red{60.76} & \red{43.64} & 88.62 & \red{87.17} & 82.66 & \red{84.86} & \red{73.71} & \red{99.08}\\
\hline
\end{tabular}
\end{table*}

\begin{table*}[!t]
\vspace{-0.3cm}
\caption{Comparison of different backbone architectures on the three datasets. Color convention: \red{best}, \blue{2nd-best}, and \yellow{3rd-best}.}
\label{tab:ablation_study_2}
\centering
\renewcommand{\arraystretch}{1.2} 
\setlength{\tabcolsep}{3pt} 
\begin{tabular}{l|@{\hspace{1.5pt}}c@{\hspace{2.5pt}}c@{\hspace{1.5pt}}|ccccc|ccccc|ccccc}
\hline
\multirow{2}{*}{{Methods}} & \multirow{2}{*}{{\makecell{Params\\(M)$\downarrow$}}} & \multirow{2}{*}{{\makecell{FLOPs\\(G)$\downarrow$}}} & \multicolumn{5}{c|}{{VisNIR-HCD (\%)}} & \multicolumn{5}{c|}{{MT-Wuhan (\%)}} & \multicolumn{5}{c}{{XiongAn (\%)}} \\
\cline{4-18} 
 & & & \makebox[2.4em]{Pre $\uparrow$} & \makebox[2.4em]{Rec $\uparrow$} & \makebox[2.4em]{F1 $\uparrow$} & \makebox[2.4em]{IoU $\uparrow$} & \makebox[2.4em]{OA $\uparrow$} & \makebox[2.4em]{Pre $\uparrow$} & \makebox[2.4em]{Rec $\uparrow$} & \makebox[2.4em]{F1 $\uparrow$} & \makebox[2.4em]{IoU $\uparrow$} & \makebox[2.4em]{OA $\uparrow$} & \makebox[2.4em]{Pre $\uparrow$} & \makebox[2.4em]{Rec $\uparrow$} & \makebox[2.4em]{F1 $\uparrow$} & \makebox[2.4em]{IoU $\uparrow$} & \makebox[2.4em]{OA $\uparrow$} \\
\hline
(l) ResNet18 \cite{He2016_ResNet} & 16.35 & 21.09 & 81.14 & \red{78.71} & \yellow{79.90} & \yellow{66.53} & 98.81 & 56.59 & \red{57.86} & \yellow{57.22} & \yellow{40.07} & \yellow{87.90} & 84.77 & \red{84.20} & \yellow{84.48} & \yellow{73.14} & \yellow{99.04}\\
(m) EfficientNet \cite{Tan2019_EfficientNet} & 8.18 & 18.41 & \yellow{83.06} & \blue{77.43} & \blue{80.15} & \blue{66.87} & \blue{98.84} & \blue{61.45} & \yellow{55.41} & \blue{58.27} & \blue{41.12} & \blue{88.90} & \blue{86.92} & 82.01 & 84.39 & 73.00 & \blue{99.06} \\
(n) MobileViT \cite{Mehta2022_MobileViT} & 9.80 & 18.90 & 79.97 & 70.33 & 74.84 & 59.80 & 98.58 & 49.97 & 54.67 & 52.21 & 35.33 & 86.01 & 83.66 & 82.63 & 83.14 & 71.15 & 98.96 \\
(o) ConvNeXt \cite{Liu2022_ConvNeXt} & 34.56 & 64.48 & \blue{83.30} & 76.17 & 79.58 & 66.08 & \yellow{98.82} & \yellow{56.84} & 54.27 & 55.52 & 38.43 & 87.84 & \yellow{86.54} & \yellow{82.65} & \blue{84.55} & \blue{73.24} & \blue{99.06} \\
(p) LWGANet \cite{lu2026lwganet} & 23.67 & 21.22 & \red{84.77} & \yellow{76.46} & \red{80.40} & \red{67.23} & \red{98.88} & \red{62.73} & \blue{56.22} & \red{59.29} & \red{42.14} & \red{89.21} & \red{86.99} & \blue{83.03} & \red{84.96} & \red{73.85} & \red{99.09} \\
\hline
\end{tabular}
\vspace{-0.2cm}
\end{table*}

\subsection{Qualitative Analysis}

To visually assess and compare the detection performance among different CD approaches, we also carry out qualitative visualizations on the three datasets, which are presented in Fig. \ref{fig:visual_results}. The qualitative results presented in these figures illustrate that our method exhibits superiority over other CD methods.

In the VisNIR-HCD dataset, significant spectral reflectance differences in vegetation and building roofs between RGB and NIR bands constitute the primary source of pseudo-changes. Empirical observations reveal that methods such as AFENet~\cite{Pu2024}, STADE-CDNet~\cite{Li2024_STADE}, and ConvFormer-CD~\cite{Yang2025_ConvFormer} fail to effectively decouple these spectral discrepancies, resulting in dense FPs in vegetation-covered areas. Similarly, MambaCD~\cite{Chen2024_MambaCD} frequently misclassifies seasonal vegetation variations as building changes. In contrast, the prediction maps generated by ASFR-Net exhibit the highest purity, characterized by coherent change regions and minimal false positive noise. This superior robustness demonstrates that GADU successfully aligns heterogeneous features via modality-invariant learning, while SFEM leverages frequency-domain filtering to further suppress residual spectral deviations, thereby enabling the model to focus on genuine semantic structural changes.

For the MT-Wuhan and XiongAn datasets, the inherent SAR speckle noise often compromises model inference. STADE-CDNet~\cite{Li2024_STADE} and STENet~\cite{Pan2024_STENet} falter significantly when processing these samples, misidentifying the granular texture of SAR as changes. In the complex scenes of XiongAn, particularly in rows 10 and 11, these methods generate extensive blocky false alarms marked in red. Moreover, DGMA2-Net~\cite{Ying2024_DGMA2} and CASP~\cite{Wang2025_CASP} exhibit marked discontinuities and missed detections (FNs) when detecting the slender road in row 8, revealing the limitations of context modeling under strong noise. Conversely, ASFR-Net demonstrates excellent noise resistance and object completeness. As shown in row 7 and row 11 of the figure, it successfully suppresses background speckles that confuse other models and accurately extracts complete change shapes. This performance is attributable to the frequency-domain purification and multi-scale aggregation strategies in SFEM, ensuring that the generated detection map is internally coherent and contiguous, significantly reducing the false alarm rate while ensuring a competitive recall.

In urban monitoring tasks, accurately delineating building boundaries and separating closely adjacent targets are critical for evaluating model practicality. In the dense urban scenes of the XiongAn dataset (row 12), affected by the geometric distortion of SAR side-looking imaging, the prediction maps generated by RFANet~\cite{You2024_RFANet} and EFICNN~\cite{Liu2025_EFICNN} suffer from boundary erosion and incomplete detection, represented by FNs. HRMNet~\cite{Li2025_HRMNet} also manifests edge blurring when detecting large factories in VisNIR-HCD. In contrast, ASFR-Net exhibits superior boundary preservation in all scenarios. Even in the high-density building area of XiongAn, it generates sharp and regular contours and clearly separates adjacent targets. This precision is primarily attributed to the synergy between the deep semantic guidance of HGFM and the explicit geometric constraints of the PAFR loss. By regulating class margins in the feature space, ASFR-Net effectively suppresses prevalent blurring artifacts and ensures high structural fidelity.

Despite the superior qualitative and quantitative performance, ASFR-Net is constrained by a fundamental limitation inherent to cross-modal tasks: it cannot resolve changes that are inherently unobservable in one modality. The practical boundary for ASFR-Net to work reliably relies on the premise that a genuine semantic change must leave an observable, structure-related imprint, such as clear edges or regular geometries, in both modalities, even if their radiometric relationship is non-linear. Consequently, there are inevitable modal blind spots where the method may experience performance degradation, as evidenced by the minor error regions in Fig. \ref{fig:visual_results}. The first failure scenario is low-contrast structure invisibility, leading to missed detections (FNs). For instance, in row 3 of VisNIR-HCD, a distinct green gap appears at the edge of the large building on the right side of our prediction map. Observing the corresponding $I_{T_2}$ reveals that this roof section exhibits lower reflectance, making it visually indistinguishable from the adjacent dark ground or shadows. The second scenario is deceptive structure interference, causing false alarms (FPs). This occurs when large-scale non-building events generate strong, regular geometric patterns in a specific modality, coincidentally mimicking the structural characteristics of buildings. As observed in row 10 of Fig. \ref{fig:visual_results}, several distinct red patches appear below the correctly detected white region. Although $I_{T_1}$ shows bare land in that area, $I_{T_2}$ exhibits highly structured, bright scattering signals. These strong non-building geometric patterns deceive the network. Ultimately, ASFR-Net cannot intrinsically resolve fundamentally unobservable changes; rather, it suppresses residual modal disparities, provided that the underlying structural evidence of the targets remains intact.

\begin{figure}[!t]
\centering
\includegraphics[width=\linewidth]{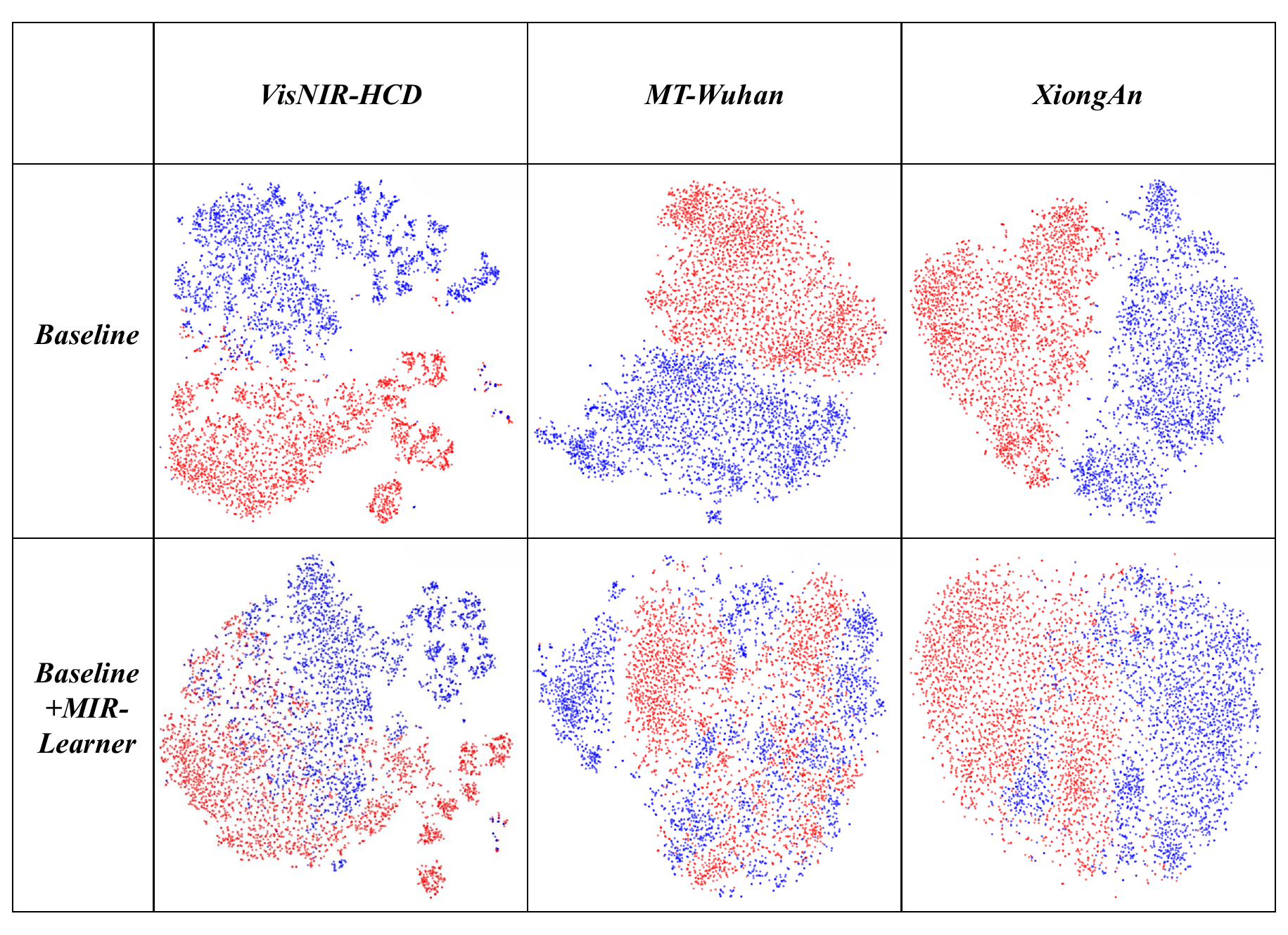}
\vspace{-0.6cm}
\caption{Comparative t-SNE visualization of feature distributions with and without the MIR-Learner. The blue and red points correspond to features from two heterogeneous modalities like optical and SAR or NIR data.}
\label{fig:tsne}
\end{figure}

\begin{table}[!t]
\vspace{-0.2cm}
\caption{Hyperparameter Sensitivity Analysis on the VisNIR-HCD Dataset. Color convention: \red{best}, \blue{2nd-best}, and \yellow{3rd-best}.}
\label{tab:hyperparameter}
\centering
\renewcommand{\arraystretch}{1.2} 
\setlength{\tabcolsep}{1.5pt} 
\begin{tabular}{@{} l | c c c c | c c c c c @{}}
\hline
Settings & $\alpha$ & $\beta$ & $\lambda_{\text{adv}}$ & $\lambda_{\text{con}}$ & Pre $\uparrow$ & Rec $\uparrow$ & F1 $\uparrow$ & IoU $\uparrow$ & OA $\uparrow$ \\
\hline

$\alpha$ too small & 0.5 & 0.05 & 0.001 & 0.002 & \red{84.46} & 76.68 & \blue{80.38} & \blue{67.20} & \red{98.87} \\
$\alpha$ too strict & 1.5 & 0.05 & 0.001 & 0.002 & \blue{84.23} & 76.46 & 80.15 & 66.88 & \blue{98.86} \\
Zero tolerance & 1.0 & 0.0 & 0.001 & 0.002 & 81.10 & \red{79.25} & \yellow{80.16} & \yellow{66.90} & 98.82 \\
$\beta$ too loose & 1.0 & 0.2 & 0.001 & 0.002 & 81.90 & \yellow{78.12} & 79.96 & 66.62 & 98.82 \\
Wrong magnitude & 1.0 & 0.05 & 0.01 & 0.02 & 83.60 & 76.72 & 80.01 & 66.68 & 98.84 \\
Ratio inverted & 1.0 & 0.05 & 0.002 & 0.001 & \yellow{84.07} & 76.19 & 79.93 & 66.57 & \yellow{98.85} \\
\textbf{Ours} & \textbf{1.0} & \textbf{0.05} & \textbf{0.001} & \textbf{0.002} & 82.45 & \blue{78.53} & \red{80.44} & \red{67.28} & \blue{98.86} \\
\hline
\end{tabular}
\vspace{-0.2cm}
\end{table}

\begin{figure*}[!t]
\centering
\includegraphics[width=\textwidth]{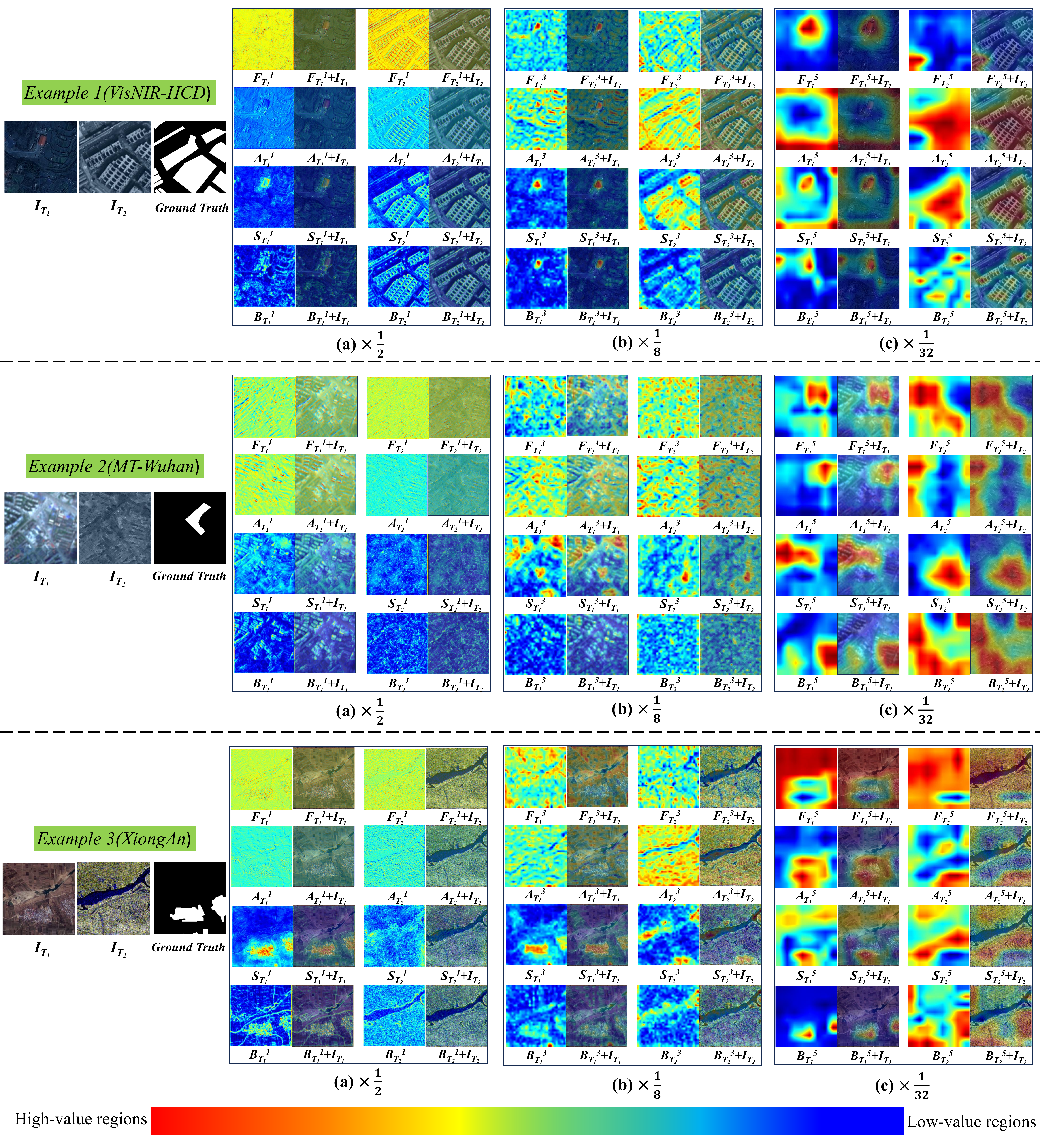}
\vspace{-0.6cm}
\caption{Progressive heatmap visualizations of multiple levels for ASFR-Net on the three datasets. $I_{T_{1}}$ and $I_{T_{2}}$ are the bi-temporal images, followed by the corresponding ground truth. (a)--(c) visualize the features from three distinct hierarchical levels, corresponding to downsampling rates of $1/2$, $1/8$, and $1/32$, respectively. Each panel sequentially displays four feature states from top to bottom: the raw features $F_{T_t}^i$ extracted by the backbone, the intermediate features $A_{T_t}^i$ purified by ASAU, the reinforced features $S_{T_t}^i$ aggregated by HPAU, and the features $B_{T_t}^i$ generated by the baseline (model (a)) in Table \ref{tab:ablation_study_1}, which utilizes basic convolutions for alignment. Each state is presented alongside its composite visualization overlaid on the original image, such as $F_{T_t}^i + I_{T_t}$. A standard jet color map is employed, where warm colors indicate high activation values and cool colors  represent low activation or suppressed regions.}
\label{fig:heap_vis}
\vspace{-0.2cm}
\end{figure*}

\subsection{Ablation Study}
\label{sec:ablation}

To verify the effectiveness and necessity of the core components in ASFR-Net, we conduct detailed ablation experiments on all three datasets. The evaluated components include MIR-Learner, SFEM, and HGFM. Furthermore, we also evaluate the performance of ASFR-Net when incorporating different backbone alternatives as substitutions. The quantitative results are summarized in Table \ref{tab:ablation_study_1} and Table \ref{tab:ablation_study_2}. The baseline (model (a)) is defined as a weight-sharing Siamese MobileNetV2 backbone with basic difference fusion and a simplified feature pyramid network decoder, without the proposed modules.

\textbf{Effectiveness of Core Components:} We first evaluate the effectiveness of MIR-Learner. Incorporating MIR-Learner in model (b) substantially improves the F1 from 77.59\% to 79.38\% on the VisNIR-HCD dataset. To isolate the specific contribution of the geometric constraint, model (c) is evaluated by removing the PAFR. This exclusion leads to a notable decline in Pre on VisNIR-HCD from 85.16\% to 81.42\%. This shows that while the adversarial component aligns global distributions, the PAFR is indispensable for maintaining feature discriminability by enforcing a margin between changed and unchanged features in the latent space. The t-SNE visualization in Fig. \ref{fig:tsne} further corroborates that MIR-Learner effectively aligns distributions while preserving semantic separability.

Building upon the aligned features, SFEM is introduced to suppress residual modal noise and recover fine-grained structural details. The integration of the full SFEM in model (d) yields significant quantitative gains, particularly in Rec which increases from 58.23\% to 63.26\% on the MT-Wuhan dataset. Removing HPAU in model (e) results in a loss of structural integrity and the exclusion of ASAU in model (f) degrades performance, suggesting that the network struggles to filter sensor-specific noise without frequency-domain analysis. These quantitative comparisons validate that ASAU and HPAU work synergistically to decouple modal artifacts from genuine changes. To intuitively comprehend this refinement process, Fig. \ref{fig:heap_vis} visualizes the multi-scale feature heatmaps for three examples. As clearly observed, the features extracted by the baseline model ($B_{T_t}^i$) suffer from severe modal interference, heavily activating on background noise such as SAR speckles or spectral deviations. In contrast, within our framework, the raw backbone features ($F_{T_t}^i$) first capture the coarse regions. After being explicitly purified by the ASAU in the frequency domain ($A_{T_t}^i$) and structurally aggregated by the HPAU ($S_{T_t}^i$), the task-irrelevant modal artifacts are effectively suppressed. Consequently, the final representations ($S_{T_t}^i$) strictly focus on genuine semantic structural changes, visually corroborating the superiority of the proposed SFEM. 

Furthermore, replacing the baseline decoder with the proposed HGFM blocks in model (g) provides consistent improvements across all metrics. For instance, on the XiongAn dataset, HGFM improves the IoU from 71.87\% to 72.54\%, demonstrating that the top-down guidance strategy effectively bridges the semantic-spatial gap by using deep semantics to refine the boundaries of shallow difference features.

The complete ASFR-Net (model k), which integrates all the aforementioned modules, achieves the best overall performance across all benchmarks. The synergistic effect is clearly demonstrated when observing the performance degradation in combinations where key components are absent. For example, combining only SFEM and HGFM in model (j) without MIR-Learner leads to a sharp drop in F1 on MT-Wuhan from 60.76\% to 59.37\%, highlighting the necessity of adversarial alignment for bridging large modal gaps. 

\textbf{Choice of Backbone Architectures:} Finally, Table \ref{tab:ablation_study_2} assesses different backbone architectures to justify our design choice. Compared to heavier CNNs, namely ResNet18 \cite{He2016_ResNet} and ConvNeXt \cite{Liu2022_ConvNeXt}, our MobileNetV2-based model achieves superior accuracy with significantly fewer parameters. Training logs further reveal that over-parameterized models suffer from validation fluctuations, likely overfitting to sensor-specific modal noise rather than learning generalizable cross-modal semantics. Among efficient alternatives, EfficientNet \cite{Tan2019_EfficientNet} delivers a highly competitive F1 score but incurs higher computational overhead. The recently proposed remote sensing backbone LWGANet \cite{lu2026lwganet} demonstrates remarkable capabilities. Although its parameter count is notably larger than that of MobileNetV2, its efficient grouped attention mechanism ensures that the actual GPU memory footprint as well as the training and inference times remain surprisingly comparable to our lightweight design. Furthermore, the final performance metrics of LWGANet are highly competitive and almost identical to our baseline model across most evaluation tracks. Nevertheless, our MobileNetV2-based architecture maintains a slight advantage on the highly noisy MT-Wuhan dataset while requiring nearly a quarter of the parameters. Consequently, considering the experimental outcomes across the three datasets, MobileNetV2 is identified as the optimal architecture for this task, as it achieves the best overall detection accuracy and superior training stability while simultaneously demanding the lowest computational overhead.

\textbf{Hyperparameter Sensitivity Analysis:} To justify the selected hyperparameter configurations, we evaluated ASFR-Net under specific boundary variations on the VisNIR-HCD dataset, as summarized in Table \ref{tab:hyperparameter}.

First, regarding the geometric boundaries ($\alpha, \beta$) in the PAFR module: In the $L_2$-normalized feature space, setting $\alpha=1.0$ guarantees an angular separation of at least $60^\circ$ between changed feature pairs. As shown in Table \ref{tab:hyperparameter}, reducing $\alpha$ to 0.5 fails to push changed features far enough apart, causing a slight drop in F1. Conversely, setting $\alpha=1.5$ imposes an overly strict constraint, disrupting the feature manifold optimization and degrading F1 to 80.15\%. For the attraction margin $\beta$, setting it to zero tolerance of 0.0 will force the network to penalize inherent sensor noise, causing modality-specific overfitting and a noticeable Pre drop. Setting $\beta=0.2$ results in loose clustering, generating excessive false positives and dropping the overall F1 to 79.96\%.

Second, regarding the gradient balancing coefficients ($\lambda_{\text{adv}}, \lambda_{\text{con}}$): These parameters dictate the multi-task optimization dynamics. When scaled up to Wrong magnitude, the massive gradients generated by the dense multi-scale adversarial and contrastive pyramids disrupt the pixel-level segmentation loss, leading to sub-optimal convergence, F1 dropping to 80.01\%. When the ratio is inverted ($\lambda_{\text{adv}}=0.002 > \lambda_{\text{con}}=0.001$), the aggressive global distribution alignment overshadows the local geometric constraints. This over-alignment inadvertently smooths out fine-grained boundary details, degrading F1 to 79.93\%. Thus, maintaining the $10^{-3}$ magnitude and with $\lambda_{\text{con}} > \lambda_{\text{adv}}$ provides a slightly stronger contrastive push, acting as the optimal gentle manifold regularizer.

\section{Conclusion and Future Work}
In this article, we propose ASFR-Net, a robust framework designed to reconcile the intrinsic conflict between feature alignment and semantic discriminability in heterogeneous CD. Adopting a coarse-to-fine refinement paradigm, the model synergizes adversarial learning with frequency-domain analysis. Specifically, MIR-Learner mitigates the primary modal discrepancy by enforcing geometric regularity, while SFEM demonstrates that residual modal discrepancies, which are often indistinguishable in the spatial domain, can be effectively decoupled and suppressed via frequency-domain priors. Then, HGFM bridges the semantic-spatial gap by employing hierarchical guided fusion, ensuring that deep semantic priors effectively sharpen fine-grained details for precise boundary delineation. Furthermore, the VisNIR-HCD benchmark alleviates the data scarcity issue in multi-source building CD. Extensive empirical evaluations substantiate that ASFR-Net achieves SOTA performance, demonstrating superior robustness against complex spectral heterogeneity and noise.

Although the current framework achieves remarkable performance, its practical application in complex real-world scenarios remains constrained by strict spatial image alignment. Future work will be dedicated to breaking this limitation, further enhancing the robustness and practicality of the model. Since perfectly aligned heterogeneous image pairs are scarce in rapid‑response scenarios due to sensor parallax, jitter, and imaging principle differences, we plan to explore a unified architecture that jointly optimizes cross‑modal registration and change detection in a mutually reinforcing manner. By investigating advanced representation disentanglement strategies, we aim to effectively separate the features required for spatial alignment from those critical for change identification. Meanwhile, we will explore task-driven feedback mechanisms, allowing the downstream CD process to actively guide and refine the spatial registration. Through this joint paradigm, we expect to minimize misalignment-induced pseudo-changes, mitigate the impact of adverse artifacts, and ultimately prevent the performance degradation of downstream semantic tasks.

\bibliographystyle{IEEEtran} 
\bibliography{ref.bib} 




\end{document}